\documentclass[preprint,12pt]{elsarticle}

\usepackage{amssymb}
\usepackage{amsmath}
\usepackage{threeparttable}
\usepackage{booktabs}
\usepackage{subfigure}
\usepackage{graphicx}
\usepackage{picins}
\usepackage{xcolor, soul}
\usepackage{url}

\journal{Information Sciences}

\begin{document}

\begin{frontmatter}



\title{Intensity-free Convolutional Temporal Point Process: Incorporating Local and Global Event Contexts}


\author[uestc]{Wang-Tao Zhou}
\ead{wtzhou@std.uestc.edu.cn}
\author[uestc]{Zhao Kang}
\ead{zkang@uestc.edu.cn}
\author[uestc]{Ling Tian\corref{mycorrespondingauthor}}
\ead{lingtian@uestc.edu.cn}
\cortext[mycorrespondingauthor]{Corresponding author}
\author[uestc]{Yi Su}
\ead{202112081359@std.uestc.edu.cn}
\affiliation[uestc]{organization={University of Electronic Science and Technology of China},
            addressline={2006 Xiyuan Avenue}, 
            city={Chengdu},
            postcode={611731}, 
            state={Sichuan},
            country={China}}

\begin{abstract}
Event prediction in the continuous-time domain is a crucial but rather difficult task. Temporal point process (TPP) learning models have shown great advantages in this area. 
Existing models mainly focus on encoding global contexts of events using techniques like recurrent neural networks (RNNs) or self-attention mechanisms. However, local event contexts also play an important role in the occurrences of events, which has been largely ignored.
Popular convolutional neural networks, which are designated for local context capturing, have never been applied to TPP modelling due to their incapability of modelling in continuous time. In this work, we propose a novel TPP modelling approach that combines local and global contexts by integrating a continuous-time convolutional event encoder with an RNN. The presented framework is flexible and scalable to handle large datasets with long sequences and complex latent patterns. The experimental result shows that the proposed model improves the performance of probabilistic sequential modelling and the accuracy of event prediction. To our best knowledge, this is the first work that applies convolutional neural networks to TPP modelling.

\end{abstract}



\begin{keyword}
Temporal Point Process \sep Convolution \sep local context \sep event prediction



\end{keyword}

\end{frontmatter}


\section{Introduction}
\label{introduction}
Events, characterized as occurrences with specific time, location and semantics, are a useful tool for describing real-world dynamics. Predicting future events can help people anticipate risks, reduce loss, and optimize welfare. For example, event prediction techniques have been applied in many areas, such as social unrest forecasting~\mbox{\citep{dynamicsocial}}, traffic accident prediction~\mbox{\citep{traffic}}, financial event prediction~\mbox{\citep{finance}}, natural disaster forecasting~\mbox{\citep{earthquake}}, recommendations~\mbox{\citep{HawkesRecommend}}, etc.

Event prediction mechanism has been extensively researched in recent years. In many scenarios, event prediction is performed in discrete time~\citep{multitask,dynamicknowledge}, where time is split into discrete windows. When it comes to time-sensitive event prediction tasks, these approaches fail to accurately predict the specific arrival time of the next event, which limits its application in practice. To achieve fine-grained event prediction in continuous time, TPPs~\citep{pointprocess} have been extensively studied. Generally speaking, TPPs describe discrete event occurrences along a continuous timeline. Each event is characterized by a mark and an arrival timestamp. Note that events may distribute non-uniformly on the timeline, which means the intervals between every two consecutive events are not equal. By learning the complicated correlations and dependencies in TPP sequences, we can predict the joint distribution of the mark and arrival time of the next event. 

\begin{figure*}[ht!]
	\centering
	\includegraphics[width=\textwidth]{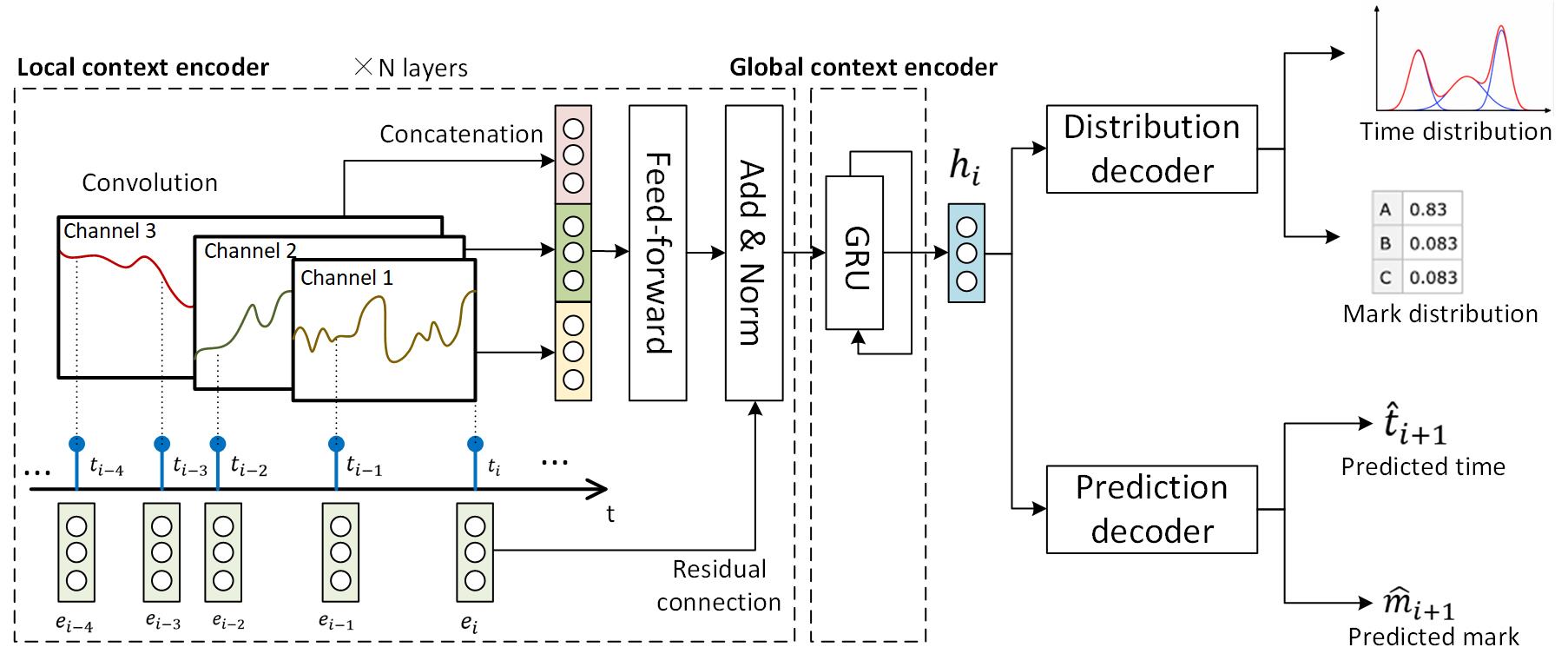}
	\vspace{-0.5em}
	\caption{Overview of the proposed CTPP framework. The local encoder on the left side consists of multiple continuous convolution kernels (channels) that aggregates event information within a certain horizon. The outputs of all kernels are concatenated and passed to a feed-forward layer. A residual connection and a layer normalization are applied. Multiple layers of local encoders can be stacked to capture more comprehensive local features. The final local features are then passed to a GRU sequentially. The GRU acts as a global feature encoder that captures long-range dependency. The hidden states can be decoded to model the probabilistic distribution of the next event or to predict the event directly.}
	\centering
	\label{fig:framework}
	\vspace{-1em}
\end{figure*}
A series of TPP models have been proposed for continuous-time event prediction. The vanilla self-exciting Hawkes process~\citep{Hawkes} is one of the classical TPP models, which assumes a parametric form of the intensity function. For better flexibility, several neural TPP models have been proposed. ~\citep{rmtpp,lstmtpp} propose to encode history events into a fixed-size vector with an RNN and generate the intensity function with the hidden state. ~\citep{nhp} replaces the history encoder with a novel continuous-time LSTM network to improve flexibility.~\citep{FullyNN} proposes to parametrize the cumulative intensity with a fully connected neural network.
~\citep{SAHP,THP,Fourier,2022transformer} adopt the Transformer structure to encode event correlation, which is better at modelling long-term dependency. ~\citep{lognormmix} proposes to generate the objective distribution in an intensity-free way, with a mixture distribution model, boosting both prediction accuracy and efficiency.~\citep{unipoint} constructs the intensity function with the sum of basis functions to increase flexibility. 

\begin{figure}
    \centering
    \includegraphics[width=\textwidth]{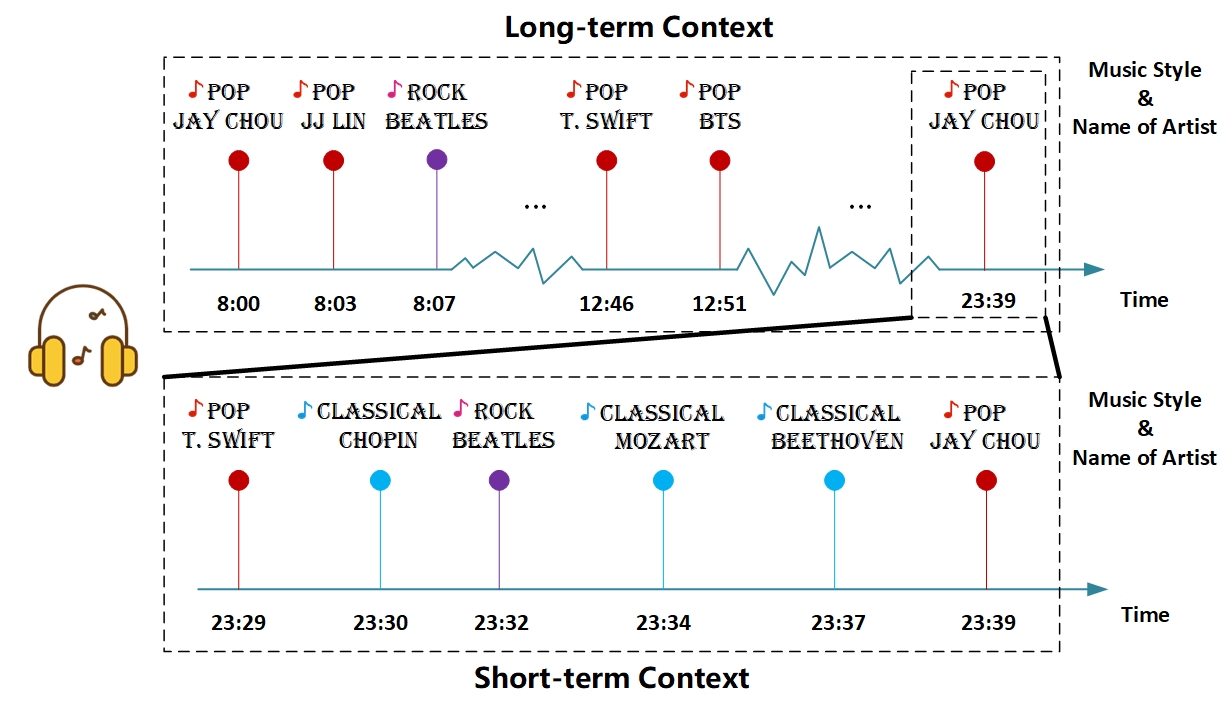}
    \caption{A concrete example based on music listening event prediction that jointly considers long and short-term event contexts. Long-term context generalizes the global tendency while short-term context focuses more on the local structure. Incorporating both of them can help us construct the event context more comprehensively.}
    \label{fig:example}
\end{figure}

An event's semantic is not only determined by itself but also by its context. The global context of an event refers to the long-term information carried by the whole event sequence ahead of it, while the short-term context refers to the information provided by recent event occurrences within a certain horizon with respect to the underlying event. Fig. \mbox{\ref{fig:example}} illustrates a concrete example of the long and short-term contexts of events that influence future event occurrences. This example is based on LastFM dataset, which contains records of songs (music style, name of artist, etc.) listened to by certain users. The long-term context of music listening of a user indicates that he likes pop music and listens to music constantly. If we are to predict his future music listening events, we would tend to infer that he would listen to another pop song, likely sung by a famous pop singer like Jay Chou, within a very short time. However, if we focus on the short-term context, we can find that he has listened to several classical music pieces within the last 10 minutes. Then, a different  prediction might be made since recent events have a greater influence on the future than events far in time. He could be trying to fall asleep or reading a book and does not want to listen to loud music that much in a short time. Thus, when predicting future events, both long and short-term contexts should be incorporated to make more accurate predictions. Similar issues have been addressed by \mbox{\citep{eg1,eg2}}, which propose to assign the users to different states (e.g., sensitization or boredom) considering their recent behaviours, in order to boost recommendation performance.  State-of-the-art neural TPP models either adopt a recurrent structure or the self-attention mechanism to utilize the global context of events for prediction purposes. However, they fail to emphasize the local context of temporal events.

As demonstrated by self-exciting event prediction models~\citep{self-exciting}, it is reasonable to assume that recent events cast greater influence on the semantics of the current event. For example, it is common sense in seismology that aftershocks can be better predicted considering information regarding recent earthquakes. Thus, capturing information induced by recent event observations helps us construct a comprehensive semantic representation of events. As proposed by~\citep{global_local}, combining the global model with local features improves the performance of time series forecasting. However, the significance of local contexts has never been emphasized in TPP modelling.

To better capture local contexts of temporal events, we introduce Convolutional Neural Networks (CNNs) into TPP modelling, which has proved to be good at capturing local structures. As a mature technique in deep learning, CNNs~\citep{convolution} have been applied in sequence modelling ~\citep{conv2}, with practical scenarios such as language modelling ~\citep{conv1,conv3}, audio generation~\citep{wavenet}, speech recognition~\citep{speech}, anomaly detection \mbox{\citep{anomaly}}, etc. \citep{TCN} introduces the Temporal Convolutional Network (TCN), which is a multi-layer dilated CNN working in discrete time.~\citep{ckconv} proposes a continuous convolution mechanism facilitating the construction of continuous-time convolutional networks.
Inspired by it, we propose an intensity-free Convolutional Temporal Point Process (CTPP) to accomplish global \& local information aggregation, so as to boost the performance of TPP modelling and continuous-time event prediction. The proposed architecture is illustrated in Fig.\ref{fig:framework}. A multi-layer local event encoder implemented with a continuous-time CNN is adopted to aggregate local event features, while a Gated Recurrent Unit (GRU)~\citep{chung2014empirical} is applied to capture global contexts. Two types of decoders can be used for intensity-free distribution modelling and event prediction, respectively.
Our contribution is three-fold:
\begin{itemize}
	\item We introduce an intensity-free continuous-time event convolutional network to capture the local context of events. This network is flexible and scalable and can focus on the local context of different horizons.
	\item We combine the event convolutional network with an RNN to fuse local and global information for better modelling performance.
	\item We conduct comprehensive experiments on three real-world datasets and compare our model with state-of-the-art models. The result shows that the proposed CTPP outperforms existing methods.
\end{itemize}

\section{Related Work}
\label{related work}
TPP modelling is an important component of time series analysis. In recent years, a sequence of neural TPP models (modelling TPPs with neural networks) has emerged. Recurrent Marked Temporal Point Process (RMTPP)~\citep{rmtpp} is one of the earliest neural TPP models. Unlike traditional parametric TPP models like Hawkes Process~\citep{Hawkes}, which sets up a fixed form of the intensity function, RMTPP proposes to parametrize the intensity with an RNN, improving flexibility and avoiding performance degradation caused by misspecification of the function form. Event history, including history event types and arrival times, is sequentially encoded to a fixed-size hidden vector, from which the intensity value at an arbitrary time can be derived. The model is trained by maximizing the likelihood of training sequences. ~\citep{lstmtpp} adopts a similar idea but uses an LSTM architecture and integrates correlated time series as input. To better capture the temporal triggering effect between events, Neural Hawkes Process (NHP)~\citep{nhp} develops a continuous-time LSTM network, which mimics the mechanism of Hawkes processes. The triggering effect is modelled as exponential decay in hidden space during time intervals. NHP does not exhibit significant performance enhancement in experiments compared to previous works, so we do not adopt the continuous-time LSTM structure in our proposed model. Instead of modelling the intensity function as a whole, UNIPoint~\citep{unipoint} proposes to generate a series of basis functions and sum them up to obtain the intensity. This approach improves the flexibility of the model and achieves better performance. Inspired by the success of Transformers in the field of natural language processing, several works~\citep{SAHP,THP,2022transformer} apply the self-attention mechanism into TPP modelling. Since self-attention is unaware of the sequential information and time intervals, a time-shifted positional encoding is incorporated.  Deep Attention Point Process (DAPP)~\citep{Fourier} uses a Fourier kernel for self-attention for better flexibility. The use of self-attention better captures long-range dependency between events, but weakens the model's ability to capture sequential dependency and hugely increases training cost due to the large number of matrix multiplications. In recent years, several generative TPP models based on variational autoencoders or denoising diffusion methods have also been derived~\mbox{\citep{generative2022,generative2020}}.

The above works model the intensity function of TPP, which makes the training process time-consuming because they have to apply the Monte-Carlo sampling technique to approximate the likelihood. To tackle this issue,~\citep{FullyNN} models the cumulative intensity function instead to avoid the intractable integral. Log Normal Mixture (LogNormMix)~\citep{lognormmix} fits the target probability density over time with a mixture distribution. Thus, the likelihood can be computed in closed form, making the optimization much more efficient. Furthermore, these methods focus on the encoding of global history information. However, sometimes the local temporal context of events has a big impact on future events, which has been largely ignored. To tackle this problem, in this work, we incorporate both local and global event contexts to obtain better performance.

Long and short-range context modelling has become a major concern in the field of Natural Language Processing (NLP) in recent years. Early language models mostly only care about short-range dependency within a window of a certain size. One typical example of such models is the N-gram model \mbox{\citep{ngram1}}. In order to capture long linguistic dependencies in a large corpus, more advanced long-range sequential networks like Transformers \mbox{\citep{bert}} are later applied to language modelling for better performance.  \mbox{\citep{globallocal1}} presents a global-local mutual attention structure, which combines an LSTM with a local short-term convolution module for long and short-term semantic feature extraction.  \mbox{\citep{globallocal2}} is a recent effort to fuse global and local semantic dependency for NLP. They propose to perform global attention and local attention separately, and then fuse them into one attention score. However, these models work on discrete words and documents, and cannot be directly applied to continuous-time modelling.

Long short-term temporal modelling has also been extensively studied in the area of recommendation systems. Incorporating long-term sequential information (global contexts) and short-term local context information is of great significance in the field of recommendation systems, because users' general tastes and recent behaviours have to be modelled jointly to provide better recommendations. \mbox{\citep{POI}} leverages an LSTM model to learn the short-term preference of users and adopts a self-attention module to aggregate global contexts.~\mbox{\citep{YUAN2020122}} proposes an attention-based context-aware sequential recommendation model using GRU to fuse global and local contexts. ~\mbox{\citep{ZHU2022}} proposes a Dynamic Global Structure Enhanced Multi-channel Graph Neural Network (DGS-MGNN), which models user preferences within the current session sequence together with rich information from a global perspective. Though the models mentioned above have studied various ways of incorporating global and local contexts of temporal sequences, they only work on discrete sequences without considering continuous time intervals, and thus cannot be used for TPP modelling directly. \mbox{\citep{CTRec}} proposes a self-attentive continuous-time recommendation model with a Demand-aware Hawkes Process (DHP) framework, which models short-term local contexts with convolutional time-aware LSTMs and adopts the self-attention mechanism to  acquire global contexts. However, unlike the simplicity of convolutional operations, the extensive use of local LSTMs makes the model harder to train. Inspired by this, we propose to use a continuous-time convolution module in our model for local context aggregation.

\section{Preliminaries}
\label{preliminaries}
\subsection{Temporal Point Process}
TPP~\citep{pointprocess} is an important tool for describing discrete event occurrences along a continuous time axis. The realization of a marked TPP is characterized as a sequence of tuples of the event marks $m_i$ and arrival times $t_i$, i.e., $\mathcal{S}=\{(m_i,t_i)\}_{i=1}^L$, where $L$ is the length of the sequence. The marks contain the semantics of the associated events. However, marks are often defined as discrete labels (event types) for simplicity, i.e., $m_i=1,2,..,K$, where $K$ is the total number of event types.

The task of TPP modelling is to fit the probability density function (pdf) $p$ of the next event given history event occurrences, i.e., 
\begin{equation}
p(m,t|H_t;\boldsymbol{\Theta})=f_{\boldsymbol{\Theta}}(m,t;H_t)
\end{equation}
where $H_t$ is the set of all history events before time $t$, and $\boldsymbol{\Theta}$ denotes the trainable parameters of the model.
However, modelling the pdf with neural networks is difficult. Most works~\citep{nhp,THP,unipoint} turn to model the conditional intensity function $\lambda_{m}(t|H_t;\boldsymbol{\Theta})$ instead, which does not have to satisfy the constraints needed for a pdf. 

The learning of a TPP model is to maximize the model's probabilistic prediction capability for future events. In practice, the log-likelihood is used as the loss function to train the model. The log-likelihood of an event sequence $\mathcal{S}$ can be calculated as:
\begin{equation}\label{log-likelihood}
\begin{aligned}
	\mathcal{L}(\mathcal{S},\boldsymbol{\Theta})&=\sum_i \log p (m_i,t_i|H_{t_i};\boldsymbol{\Theta})\\
		&=\sum_i \log \lambda_{m_i}(t_i|H_{t_i};\boldsymbol{\Theta})-\int_0^T\lambda(t|H_{t};\boldsymbol{\Theta})dt
\end{aligned}
\end{equation}
where $T$ is the upper bound of the observation time range and $\lambda(t|H_t;\boldsymbol{\Theta})=\sum_{k=1}^K \lambda_k(t|H_t;\boldsymbol{\Theta})$ is the total intensity of all event marks. The learning process of a TPP is maximizing the log-likelihood of training event sequences, or minimizing the negative log-likelihood (NLL). 

Although the intensity-based method makes it easy to model with neural networks, the computation cost induced by the intractable integral is the main drawback. Following~\citep{lognormmix}, our proposed intensity-free framework models the pdf of events directly and calculates the log-likelihood with the first line given in Eq.(\ref{log-likelihood}), avoiding the large computation cost caused by the integration term in the intensity-based version.

\subsection{Causal Convolution}
Firstly proposed by~\citep{wavenet}, the causal convolution upon temporal sequential data is a useful tool for capturing the local context of data points. The operation is performed between a convolution kernel $\boldsymbol{\psi}=\{\psi(\tau)\}_{\tau=1}^{\eta}$ and the sequential data points $\boldsymbol{x}=\{x(t)\}_{t=1}^L$ as:
\begin{equation}
(\boldsymbol{x}*\boldsymbol{\psi})(t)=\sum_{\tau=0}^{\eta}\psi(\tau)x(t-\tau)
\end{equation}
where $\eta$ stands for the size of the convolution kernel. Note that the causal convolution only aggregates past information of a data point and ignores future information. Causal temporal convolution has been widely used in synchronous time series modelling scenarios. However, traditional causal temporal convolution is unable to operate on asynchronous data, such as TPP sequences. ~\citep{ckconv} proposes to construct continuous convolution kernels with neural networks, which facilitates convolution operations for non-uniformly distributed data.
In this work, we adopt the continuous causal convolution approach to construct a convolutional local event encoder to capture local temporal contexts.

\section{Intensity-Free Convolutional Temporal Point Process}
\label{method}
In this section, we introduce the proposed CTPP model, which integrates convolution networks and RNN to incorporate local and global structures of temporal events. 

\subsection{Convolution Kernel}
\label{omega}
Unlike discrete convolution networks, where kernels can be learned as a set of discrete weights, under the setting of TPP, the convolution kernel must be a continuous function defined on the time domain. However, simple parametric functions are unable to capture complex local structures of events. We adopt a SIREN network~\mbox{\citep{siren,ckconv}} to parametrize continuous convolution kernels to facilitate flexible modelling of local temporal structures. 

The kernel is defined as:
\begin{equation}
\label{eq:omega_tau}
\boldsymbol{\psi}=\psi_{\boldsymbol{\theta}}(\tau)=\mathrm{SIREN}_{\boldsymbol{\theta}}(\tau)
\end{equation}

where $\mathrm{SIREN}$ is a multi-layer perceptron with sinusoidal activations instead of frequently used ReLU, Sigmoid, etc. $\boldsymbol{\theta}$ represents the trainable parameters associated with the kernel network. Note that $\tau$ is a continuous variable in this case. Each hidden layer of a SIREN follows the form:
\begin{equation}
y=\sin(\omega_0 (\boldsymbol{W}\boldsymbol{x}+\boldsymbol{b}))
\end{equation}
where $\boldsymbol{W}$ and $\boldsymbol{b}$ are the trainable weight matrix and bias, respectively. The periodical activation function enables the network to fit complicated functions, making it qualified to construct a continuous convolution kernel. $\omega_0$ is a hyperparameter governing the oscillation rate of the convolution kernel. We find the performance of the model is sensitive to $\omega_0$. Thus, the influence of this hyperparameter is discussed in section \mbox{\ref{param}}.

\subsection{Convolutional Local Encoder}
The convolutional local encoder is comprised of  a multi-channel convolution module and a feed-forward layer. The convolution module is defined as:
\begin{equation}
\boldsymbol{c}_i^{l}=\sum_{j;0\le t_i-t_j\le \eta^{l}} \psi^{l}_{\boldsymbol{\theta}}(t_i-t_j) \boldsymbol{e}_j
\end{equation}
where $\boldsymbol{c}_i^l$ denotes the local context feature of the $i$-th event produced by the $l$-th channel, function $\psi_\theta^{l}(\tau)$ is the continuous convolution kernel of the $l$-th channel, $\boldsymbol{e}_j \in \mathbb{R}^d$ stands for the embedding of the $j$-th event, and $\eta^{l}$ is the horizon of the $l$-th channel.  The local context encoder aggregates information of history events within a horizon to obtain the local context of an event. Note that the horizons can be set to arbitrary lengths, even infinite. We suggest tuning the horizon sizes proportional to the time scale of the dataset, using the average inter-event time $\delta$ as a basis (Table \mbox{\ref{tab:time}}). However, kernels of different horizons are able to capture the temporal dependency of different ranges. Similar to attention heads used in Transformers~\mbox{\citep{Transformer}}, different channels in the convolution layer are independent convolution kernels focusing on different patterns of local dependency. We find that the performance of the model is sensitive to the combination of the horizon size and the number of channels, which will be discussed in detail in subsecion~\mbox{\ref{param}}.
The local context features produced by the $C$ convolution channels are concatenated before passing to the feed-forward layer, and a residual connection is included for better preservation of original event semantics:
\begin{equation}
\boldsymbol{c}_i=[\boldsymbol{c}_i^1,\boldsymbol{c}_i^2,\cdots,\boldsymbol{c}_i^C]\boldsymbol{W}^O+\boldsymbol{e}_i
\end{equation}
where $\boldsymbol{W}^O\in \mathbb{R}^{Cd\times d}$ is a learnable aggregation weight matrix. Furthermore, we stack $N$ convolutional local encoders to comprehensively capture local contexts. Inspired by~\citep{Transformer}, a layer normalization is applied for better performance.

\subsection{Recurrent Global Encoder}
A GRU~\citep{chung2014empirical} network is utilized to learn global features of events. The GRU with trainable parameter $\boldsymbol{\gamma}$ takes three inputs, the last hidden state $\boldsymbol{h}_{i-1}$, the local context of the current event $\boldsymbol{c}_i$, and the time interval since the last event $t_i-t_{i-1}$, i.e.,
\begin{equation}
\boldsymbol{h}_i=\mathrm{GRU}_{\boldsymbol{\gamma}}(\boldsymbol{h}_{i-1};\boldsymbol{c}_i;t_i-t_{i-1}).
\end{equation}

The hidden state $\boldsymbol{h}_i$ summarizes local features of the first $i$ events, thus integrating local and global information of the event sequence.

\subsection{Distribution Decoder}
The target of MTPP is to model the joint probability distribution of the mark and arrival time of the next event. For simplicity, we make the assumption that the event mark and arrival time are two independent variables, leading to:
\begin{equation}
	p(m_{i+1},t_{i+1}|H_{t_i})=p(m_{i+1}|H_{t_i})p(t_{i+1}|H_{t_i})
\end{equation}
where $p(m_{i+1}|H_{t_i})$ is a categorical distribution, defined as:
\begin{equation}
	p(m_{i+1}|H_{t_i})\sim \mathrm{Categorical}(\boldsymbol{\pi}_i)
\end{equation}
\begin{equation}\label{eq:pi}
	\boldsymbol{\pi}_i=\boldsymbol{W}_\pi \boldsymbol{h}_i
\end{equation}

Following ~\citep{lognormmix}, we define the inter-event time distribution using a Log-normal mixture distribution with parameters $\boldsymbol{w}_i$, $\boldsymbol{\sigma}_i$, and $\boldsymbol{\mu}_i$, which are generated through feed-forward layers:
\begin{align}
\boldsymbol{w}_i&=\mathrm{softmax}(\boldsymbol{W}_w\boldsymbol{h}_i+\boldsymbol{b}_w)\\
\boldsymbol{\sigma}_i&=\mathrm{exp}(\boldsymbol{W}_s\boldsymbol{h}_i+\boldsymbol{b}_s)\\
\boldsymbol{\mu}_i&=\boldsymbol{W}_\mu \boldsymbol{h}_i+\boldsymbol{b}_\mu
\end{align}
Thus $\boldsymbol{\Phi}=\{\boldsymbol{W}_\pi,\boldsymbol{W}_w,\boldsymbol{W}_s,\boldsymbol{W}_\mu,\boldsymbol{b}_w,\boldsymbol{b}_s,\boldsymbol{b}_\mu\}$ are trainable parameters.
The arrival time of the next event is modelled as the following distribution:
\begin{equation}
	z_i\sim \mathrm{Categorical}(\boldsymbol{w}_i)
\end{equation}
\begin{equation}
	r_i\sim \mathcal{N}(\boldsymbol{\mu}_{i,z_i},\boldsymbol{\sigma}_i)
\end{equation}
\begin{equation}
	\hat{t}_{i+1}=\mathrm{exp}(r_i)+t_i
\end{equation}
The model integrates local contexts and global history information, making the encoded hidden states more informative, thus enabling more accurate probabilistic modelling.

\subsection{Prediction Decoder}
Distributional modelling gives the probability distribution of the mark and time of the next event. It is robust to noise, less prone to overfitting, and is capable of obtaining comprehensive statistics (expectation, variance, etc.) of the given result. However, sometimes we only expect the model to predict the exact type and arrival time of events, where probabilistic modelling becomes redundant. Furthermore, we observe that using the expected values of the mixture distribution as predictions yield poor accuracy and stability. Inspired by~\citep{THP}, we propose to predict the event type and arrival time of the next event directly from the hidden state $\boldsymbol{h}_i$. But different from~\citep{THP}, which trains a unified model that jointly optimizes probabilistic likelihood and prediction error, we propose to separate event prediction from distribution modelling as another model for better accuracy.
For event type prediction, we output the mark with the largest prediction score, defined as:
\begin{equation}
	\hat{m}_{i+1}=\underset{k}{\operatorname{argmax}}~\boldsymbol{\pi}_{ik}
\end{equation}
where $\boldsymbol{\pi}_{i}$ is calculated as Eq.(\ref{eq:pi}). The arrival time is predicted using a feed-forward layer with an exponential activation as:
\begin{equation}
	\hat{t}_{i+1}=\mathrm{exp}(\boldsymbol{W}_t\boldsymbol{h}_i+\boldsymbol{b}_t)+t_i
\end{equation}
The event prediction model with decoding parameter $\boldsymbol{\Xi}=\{\boldsymbol{W}_\pi,\boldsymbol{W}_t,\boldsymbol{b}_t\}$ is trained with a different objective to the probabilistic model, which minimizes the prediction error of the result.

\subsection{Learning}
Following previous works~\mbox{\citep{lognormmix,rmtpp,nhp}}, we propose to train our probabilistic model with Maximum Log-likelihood Estimation (MLE). Given a TPP event sequence $\mathcal{S}$, the log-likelihood loss $\mathcal{L}_{prob}$ is defined as the first line of Eq.(\ref{log-likelihood}), where the parameters $\boldsymbol{\Theta}_{prob}=\{\boldsymbol{\theta},\boldsymbol{\gamma}, \boldsymbol{W}^O,\boldsymbol{\Phi}\}$.
The training process of the model is generalized as an optimization problem:
\begin{equation}
\label{eq:argmin1}
	\boldsymbol{\Theta}^*_{prob}=\underset{\boldsymbol{\Theta}_{prob}}{\operatorname{argmin}}-\mathcal{L}_{prob}(\mathcal{S},\boldsymbol{\Theta}_{prob})
\end{equation}
This problem can be solved using existing optimization algorithms such as Adam~\citep{Adam}.

For the event prediction model, inspired by~\mbox{\citep{THP}}, we use a different objective function, which jointly evaluates the prediction loss of event marks and event times with a balancing hyperparameter $\beta$.
\begin{equation}
	\mathcal{L}_{pred}(\mathcal{S},\boldsymbol{\Theta}_{pred})=\mathcal{L}_{type}(\mathcal{S},\boldsymbol{\Theta}_{pred})+\beta\mathcal{L}_{time}(\mathcal{S},\boldsymbol{\Theta}_{pred})
\end{equation}
The type prediction loss is evaluated using the cross-entropy:
\begin{equation}
	\mathcal{L}_{type}(\mathcal{S},\boldsymbol{\Theta}_{pred})=-\sum_i \log p(m_{i+1}|H_i)
\end{equation}
Since time is a continuous variable, we evaluate the prediction loss as the sum of the squared error:
\begin{equation}
	\mathcal{L}_{time}(\mathcal{S},\boldsymbol{\Theta}_{pred})=\sum_i (t_i-\hat{t}_i)^2
\end{equation}
where $\hat{t}_i$ is the predicted time of the $i$-th event. Similar to Eq.(\ref{eq:argmin1}), the parameters $\boldsymbol{\Theta}_{pred}=\{\boldsymbol{\theta},\boldsymbol{\gamma},\boldsymbol{W}^O,\boldsymbol{\Xi}\}$ of event prediction model is learned as:
\begin{equation}
	\boldsymbol{\Theta}^*_{pred}=\underset{\boldsymbol{\Theta}_{pred}}{\operatorname{argmin}}-\mathcal{L}_{pred}(\mathcal{S},\boldsymbol{\Theta}_{pred})
\end{equation}
In our experiments, the probabilistic model and the event prediction model are separately trained and evaluated.

\section{Experimental Results and Discussion}
\label{experiments}
We evaluate our CTPP model using negative log-likelihood loss (NLL) and prediction accuracy. Experiments are conducted on three real-world datasets. The performance comparison against five baseline methods is given.

\subsection{Training Details}
We implement the proposed framework using PyTorch, and the code is available online\footnote{\url{https://github.com/AnthonyChouGit/Convolutional-Temporal-Point-Process}}. The implemented SIREN network contains three hidden layers of size 32. The local encoder consists of one or two convolution layers, and each layer contains no more than 3 channels. 
We tune the size of event embeddings and RNN hidden states in the range of [32, 64, 128]. The balancing parameter $\beta$ is tuned within [0.1, 0.3, 0.5, 0.7, 0.9]. 
The tuning strategy of the horizon size, number of channels, and oscillation rate $\omega_0$ is discussed in detail in section \mbox{\ref{param}}.
Multiple combinations of hyperparameters are tested to achieve better performance. The learning rate is set to 1e-3 initially and gradually decays along the process of training. Early stopping is applied to mitigate overfitting.

\subsection{Datasets}
\label{data}
\begin{figure}[h!]
	\centering

\subfigure[SO]{
\centering
\includegraphics[width=2.8in]{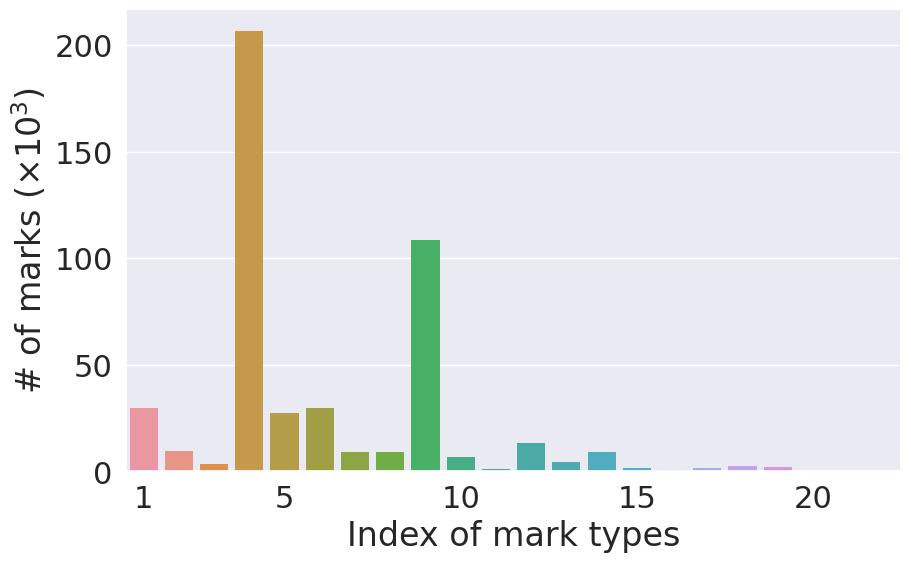}
\label{type_so}
}%
\subfigure[Retweet]{
\centering
\includegraphics[width=2in]{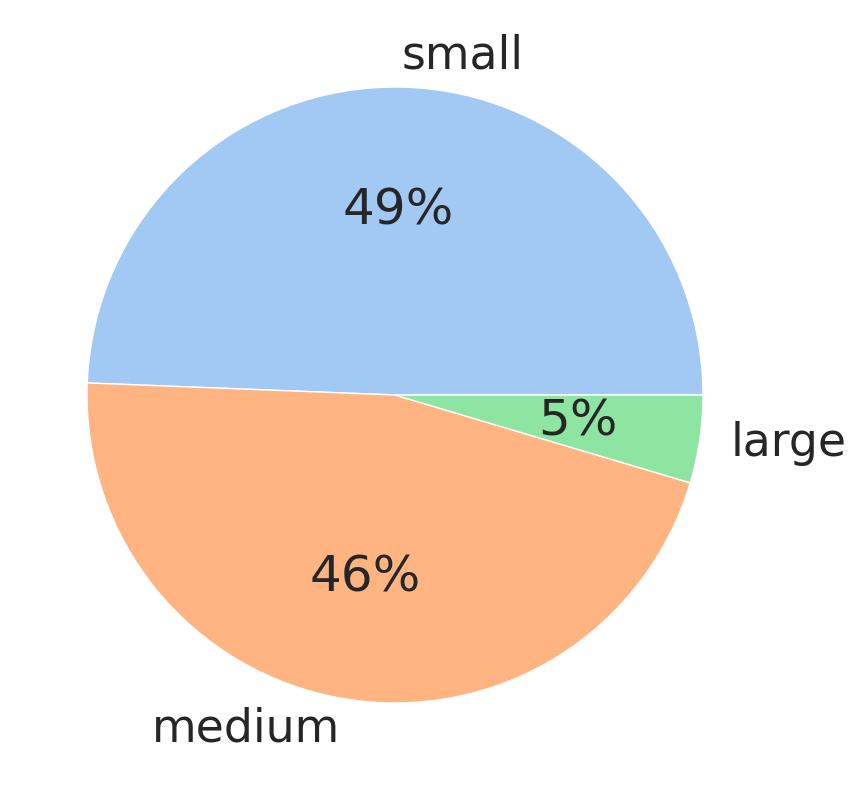}
\label{type_retweet}
}%

\subfigure[LastFM]{
\centering
\includegraphics[width=5.3in]{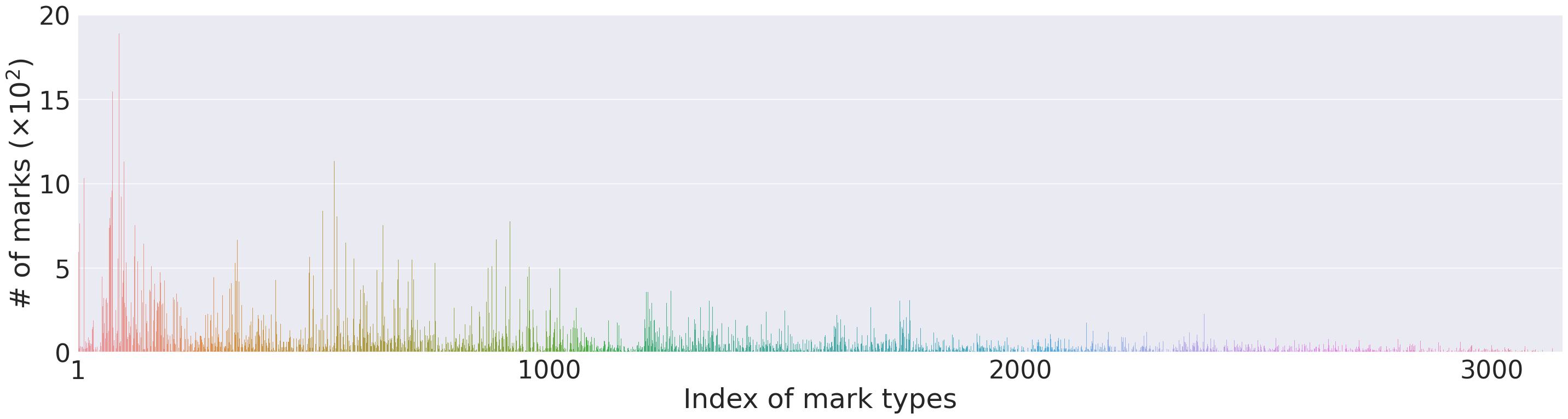}
\label{type_lastfm}
}%
%
    \caption{Mark type distributions of the datasets.}
    \label{fig:types}
\end{figure}

We use three real-world datasets, namely StackOverflow (SO), Retweet, and LastFM. See Table \ref{tbl:dataset} for detailed statistics of these datasets, where "\# of" means "number of". Table \mbox{\ref{tab:time}} is the statistics of the time scale of each dataset, which is essential for tuning hyperparameters of the convolutional model (see section \mbox{\ref{param}}). Fig.\mbox{\ref{fig:types}} exhibits the mark type distributions of each dataset. Below is a brief introduction of these datasets. 

\begin{table*}[h!]
	\centering
	\vspace{-0.5em}
	\caption{Statistics of the used datasets.}
	\vspace{0.5em}
	\begin{threeparttable}
		\begin{tabular}{@{}lrrrrrrrrr@{}}
			\hline\hline
			Dataset       & \multicolumn{1}{c}{\# of Types} & \multicolumn{3}{c}{Sequence Length}                                          & \multicolumn{3}{c}{\# of Sequences}                                               \\ \cmidrule(l){3-8} 
			& \multicolumn{1}{l}{}              & \multicolumn{1}{c}{Min} & \multicolumn{1}{c}{Mean} & \multicolumn{1}{c}{Max} & \multicolumn{1}{c}{Train} & \multicolumn{1}{c}{Validation} & \multicolumn{1}{c}{Test} \\ \midrule
			SO & 22                                & 41                      & 72                       & 256                     & 4,777 & 530 & 530      \\
			Retweet      & 3                                 & 50                      & 109                      & 256                     & 20,000                 & 2,000                        & 2,000       \\
			LastFM     & 3,150                             & 3                      & 222                        & 256                      & 743 & 93 & 93       \\
			
			\hline\hline
		\end{tabular}

	\end{threeparttable}
	\label{tbl:dataset}
	\vspace{-1em}
\end{table*}

\begin{table*}[h!]
	\centering
	\vspace{-0.5em}
	\caption{Statistics of time intervals of each dataset.}
	\vspace{0.5em}
	\begin{threeparttable}
		\begin{tabular}{@{}lrrrrrrr@{}}
			\hline\hline
			Dataset     & \multicolumn{3}{c}{Time Intervals}                                           \\ \cmidrule(l){2-4} 
			             & \multicolumn{1}{c}{Min} & \multicolumn{1}{c}{Mean($\delta$)} & \multicolumn{1}{c}{Max} \\ \midrule
			SO                                 & 0                      & 0.83                       & 20.34                     \\
			Retweet                                  & 0                      & 2583                      & 582928                      \\
			LastFM                            & 0                    & 0.76                        & 1016                      \\
			
			\hline\hline
		\end{tabular}
	\end{threeparttable}
	\label{tbl:data_time}
	\vspace{-1em}
\end{table*}

\paragraph{SO~\citep{stackoverflow}}
SO is a dataset containing records of users obtaining badges from the website Stack Overflow, which is a forum for programmers. Since there are 22 types of badges available, the types are used as the mark of each event. The type distribution is illustrated as a bar graph in Fig.\mbox{\ref{type_so}}. This dataset has generally short inter-event intervals with an average of $\delta=0.83$. 

\paragraph{Retweet~\citep{retweet}}
This dataset records the timestamps of retweets on Twitter within a certain period of time. The marks of the events denote the types of users who retweet. Specifically, users are categorized into "small", "medium" and "large" users depending on the number of their followers. The mark distribution is displayed in Fig.\mbox{\ref{type_retweet}} as a pie graph. The inter-event intervals of this dataset are of a larger time scale, with an average of $\delta=2583$.

\paragraph{LastFM~\citep{lastfm}}
LastFM is a music-related dataset, which contains records of songs listened to by certain users. The events are marked by the singer of the song. This dataset contains over 3000 mark types, whose distribution is shown as a bar graph in Fig.\mbox{\ref{type_lastfm}}. This dataset has a time scale similar to SO, with an average interval of $\delta=0.76$.

For efficient evaluation, all sequences are cut with a maximum length of 256. 

\subsection{Baselines}
We compare our proposed approach against five baseline models listed as follows.

\paragraph{Hawkes Process (HP)}
HP is a classical TPP modelling approach first proposed by~\citep{Hawkes}, which formulates the intensity function as a parametric function with learnable parameters.

\paragraph{Recurrent Marked Temporal Point Process (RMTPP)}
Proposed by~\citep{rmtpp}, RMTPP is one of the earliest works that models a TPP using RNNs, which generate the intensity values based on the hidden states of the neural network.

\paragraph{Neural Hawkes Process (NHP)}
This work~\citep{nhp} proposes a continuous-time LSTM network, which models the time intervals with exponential decaying functions.

\paragraph{Log Normal Mixture (LogNormMix)}
LogNormMix~\citep{lognormmix} is one of the state-of-the-art methods for TPP modelling, which discards the notion of intensity and concatenates a mixture distribution model with the traditional RNN encoder to model the inter-event times.

\paragraph{UNIPoint}
UNIPoint is proposed by~\citep{unipoint}. This model composes the intensity function with basis functions. The components are summed up to obtain the intensity value at a certain point. We tested UNIPoint with different kernels in our experiments and choose the one with the best performance.

\subsection{Results}
\subsubsection{NLL Test}
We perform the NLL test on three datasets and compare our proposed model with four baseline methods. The NLL results are shown in Table~\ref{tab:performance}. The result shows that CTPP has the best performance on all three datasets, which indicates that our global \& local information aggregation approach learns the history more comprehensively.
\begin{table}[!ht]
\centering
\vspace{-1em}
\caption{Negative log-likelihood per event.}
\label{tab:performance}
\begin{tabular}{l|rrrr}
\hline\hline
Dataset & \multicolumn{1}{c}{SO} & \multicolumn{1}{c}{Retweet} & \multicolumn{1}{c}{LastFM} \\ \hline
HP & 2.35 & 6.79 & 7.57  \\
RMTPP & 2.20 & 6.67 & 6.63 \\
NHP & 2.20 & 6.48 & 8.01  \\
LogNormMix & 2.03 & 3.54 & 4.73  \\
CTPP & \textbf{2.00} & \textbf{2.78} & \textbf{4.43} \\
\hline\hline
\end{tabular}
\vspace{-1em}
\end{table}

The traditional HP model performs the worst because of the lack of flexibility. The intensity-based models, RMTPP and NHP, increase the NLL result due to the application of an RNN-based flexible global encoder. LogNormMix beats previous models by adopting a flexible mixture distribution mechanism.
Note that LogNormMix has the most similar architecture as our model, but it only captures global history information, ignoring local contexts. Thus, the margin between our proposed CTPP with respect to LogNormMix proves that incorporating local temporal contexts helps us enhance TPP modelling.

Since CTPP models the mark distribution and the inter-event time distribution independently, we also compare our model with two of the baseline models in terms of mark NLL (see Table~\ref{tab:mark}) and time NLL (see Table~\ref{tab:time}) separately. HP and NHP are excluded from this comparison because these two models do not model the mark distribution and time distribution separately. UNIPoint only works on inter-event times, so it only appears in time modelling experiments.
\begin{table}[!ht]
\centering
\vspace{-1em}
\caption{Mark NLL per event.}
\label{tab:mark}
\begin{tabular}{l|rrrr}
\hline\hline
Dataset & \multicolumn{1}{c}{SO} & \multicolumn{1}{c}{Retweet} & \multicolumn{1}{c}{LastFM} \\ \hline
RMTPP & \textbf{1.52} & \textbf{0.78} & 7.34 \\
LogNormMix & \textbf{1.52} & \textbf{0.78} & 6.80  \\
CTPP & \textbf{1.52} & 0.79 & \textbf{6.59} \\
\hline\hline
\end{tabular}
\vspace{-1em}
\end{table}

As we can see from Table~\mbox{\ref{tab:mark}}, mark NLL results of the three models on StackOverflow are almost identical, and CTPP falls a little behind the other two on Retweet. This is probably a dataset-related issue. The numbers of mark types of these two datasets are rather small (see Table~\mbox{\ref{tbl:dataset}}). Additionally, the classification of the marks of Retweet is quite ambiguous (see section~\mbox{\ref{data}}). These factors might have contributed to the bottleneck of performance enhancement. However, for LastFM, which has a large number of mark types, the performance improvement of mark NLL is quite significant.
 
 Table \ref{tab:time} shows that performance improvement in terms of time NLL is consistently significant. CTPP performs better than RMTPP, LogNormMix and UNIPoint with a rather large margin on all three datasets.

\begin{table}[!ht]
\centering
\vspace{-1em}
\caption{Time NLL per event.}
\label{tab:time}
\begin{tabular}{l|rrrr}
\hline\hline
Dataset & \multicolumn{1}{c}{SO} & \multicolumn{1}{c}{Retweet} & \multicolumn{1}{c}{LastFM} \\ \hline
RMTPP & 0.67 & 5.90 & -0.72 \\
UNIPoint & 0.65 & 5.15 & -2.00 \\
LogNormMix & 0.51 & 2.76 & -2.07  \\
CTPP & \textbf{0.48} & \textbf{1.99} & \textbf{-2.16} \\
\hline\hline
\end{tabular}
\vspace{-1em}
\end{table}

\subsubsection{Event Prediction Test}
We perform event prediction tests on four models, among which "CTPP w/o local" denotes the version of our event prediction model without the local convolutional encoder. The structure of this model is similar to LogNormMix, but predicts event marks and event times directly from the hidden state rather than from the mixture distribution. For intensity-based baselines, RMTPP and NHP, the prediction is generated from the expectation of the output mark and time distribution. We evaluate the performance of event mark prediction using the prediction accuracy, see Table~\ref{tab:type acc}. The prediction result of event times is evaluated using Root Mean Squared Error (RMSE), see Table~\ref{tab:time rmse}.
\begin{table}[!ht]
\centering
\vspace{-1em}
\caption{Type prediction accuracy.}
\label{tab:type acc}
\begin{tabular}{l|rrrr}
\hline\hline
Dataset & \multicolumn{1}{c}{SO} & \multicolumn{1}{c}{Retweet} & \multicolumn{1}{c}{LastFM} \\ \hline
RMTPP & 46.35 & 44.89 & 0.82 \\
NHP & 44.89 & 46.35 & 0.86  \\
CTPP w/o local & 46.33 & \textbf{61.21} & 3.64  \\
CTPP & \textbf{46.46} & 61.16 & \textbf{3.79} \\
\hline\hline
\end{tabular}
\vspace{-1em}
\end{table}

From Table~\ref{tab:type acc}, we can find that CTPP has the best result in mark prediction on two out of the three datasets, while falling a little behind the version without the local context encoder on the Retweet dataset. Compared to intensity-based prediction, RMTPP and NHP, the performance improvement is still significant.

\begin{table}[!ht]
\centering
\vspace{-1em}
\caption{Time prediction RMSE.}
\label{tab:time rmse}
\begin{tabular}{l|rrrr}
\hline\hline
Dataset & \multicolumn{1}{c}{SO} & \multicolumn{1}{c}{Retweet} & \multicolumn{1}{c}{LastFM} \\ \hline
RMTPP & 1.26 & 16600.85 & 6.52 \\
NHP & 1.03 & 16600.85 & 6.54  \\
UNIPoint & 1.00 & 16600.32 & 6.21\\
CTPP w/o local & \textbf{0.97} & 14396.91 & 6.20  \\
CTPP & \textbf{0.97} & \textbf{14345.50} & \textbf{6.13} \\
\hline\hline
\end{tabular}
\vspace{-1em}
\end{table}

Table~\ref{tab:time rmse} shows that our model performs better than all baseline models in the time prediction test. Especially, compared to the version without the local context encoder, the proposed model with a convolutional local encoder exhibits improvements, implying that capturing local context helps us enhance model performance.

\subsection{Parameter Analysis}
\label{param}
\begin{figure}[h!]
	\centering

\subfigure[LastFM]{
\centering
\includegraphics[width=2.2in]{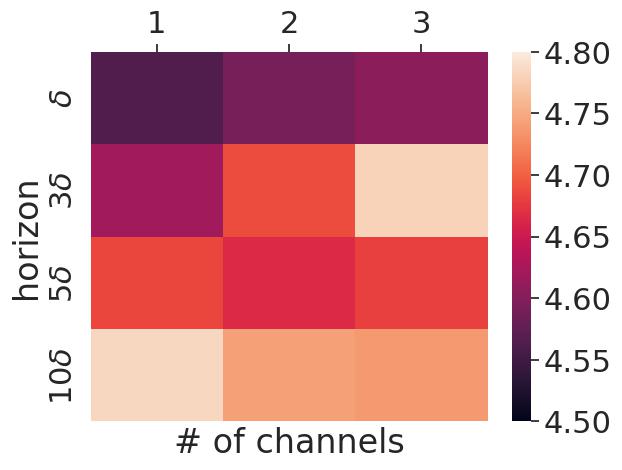}
\label{param1_a}
}%
\subfigure[Retweet]{
\centering
\includegraphics[width=2.2in]{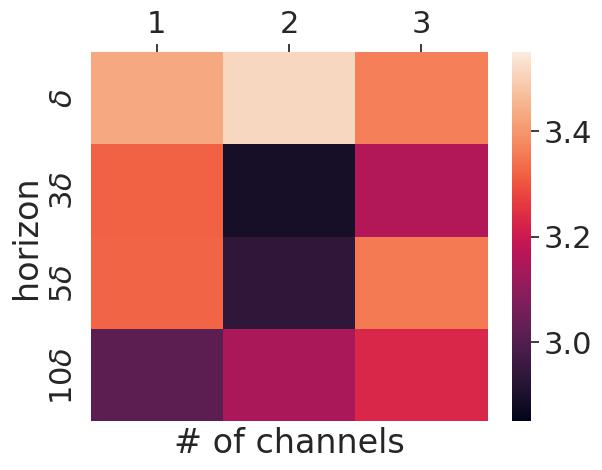}
\label{param1_b}
}%

    \caption{Parameter sensitivity analysis result of the combination of horizon size and the number of channels. The deeper the colour, the better the performance.}
    \label{fig:param1}
\end{figure}

During the experiment, we find that the experimental results are sensitive to the combination of kernel widths (referred to as horizons in previous sections), number of channels and the oscillation rate $\omega_0$. So in this section, we analyze the sensitivity of these three parameters on datasets Retweet and LastFM with the NLL test.

Fig.\mbox{\ref{fig:param1}} shows the NLL performance of CTPP having only one layer of the convolutional encoder with different combinations of the horizon and number of channels. We tune the number of channels within [1, 2, 3] and the size of the horizon within [$\delta$, $3\delta$, $5\delta$, $10\delta$], where $\delta$, as presented in section~\mbox{\ref{data}} stands for the average inter-event time interval of the corresponding dataset (e.g. the horizon tick $10\delta$ in Fig.\mbox{\ref{param1_a}} stands for a horizon size of $ 10\times 0.76=7.6$). As shown in Fig.\mbox{\ref{fig:param1}}, different combinations of the two parameters lead to different NLL results. For LastFM, Fig.\mbox{\ref{param1_a}} shows that a smaller size of the horizon and fewer number of channels leads to better performance, which means the semantics of events in this dataset are mostly influenced by local contexts within a short range, and the pattern of their dependency might be simple. For Retweet, however, Fig.\mbox{\ref{param1_b}} shows that a combination of 2 convolution channels and a horizon size of $3\delta$ or $5\delta$ has the best NLL results, which means that the events in this dataset tend to have relatively more complex local dependencies of longer range. In practice, in order to obtain better performance, multiple layers of convolution can be stacked, but the parameter tuning can be more complicated and tricky, which is beyond the scope of this paper.

\begin{figure}[h!]
	\centering

\subfigure[LastFM]{
\centering
\includegraphics[width=2.2in]{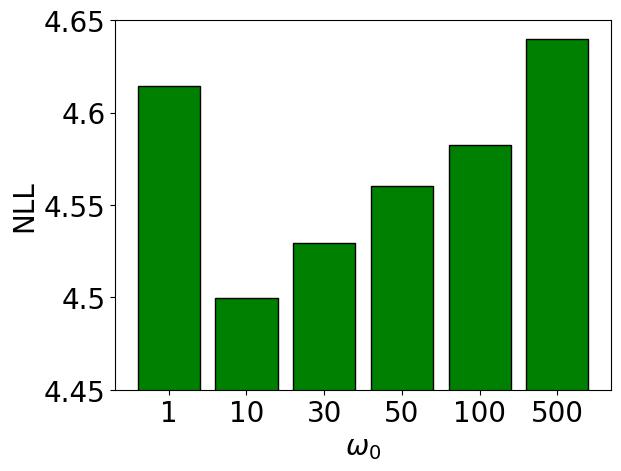}
\label{param2_a}
}%
\subfigure[Retweet]{
\centering
\includegraphics[width=2.2in]{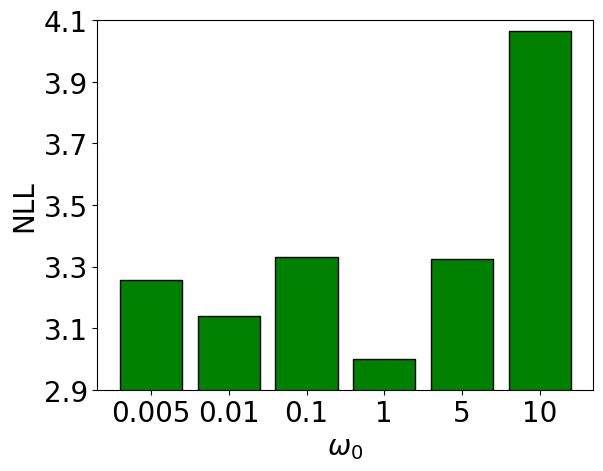}
\label{param2_b}
}%

    \caption{Performance of CTPP with different values of $\omega_0$.}
    \label{fig:param2}
\end{figure}

\begin{figure}[h!]
	\centering

\subfigure[$\omega_0=1$]{
\centering
\includegraphics[width=1.6in]{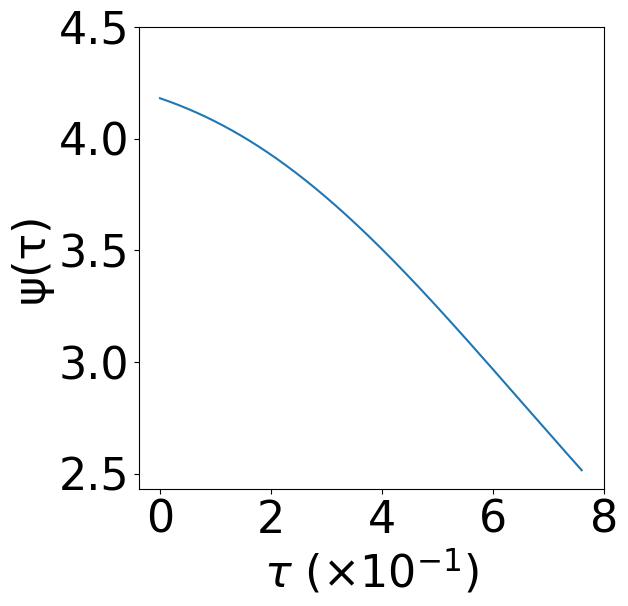}
\label{lastfm_omega1}
}%
\subfigure[$\omega_0=10$]{
\centering
\includegraphics[width=1.6in]{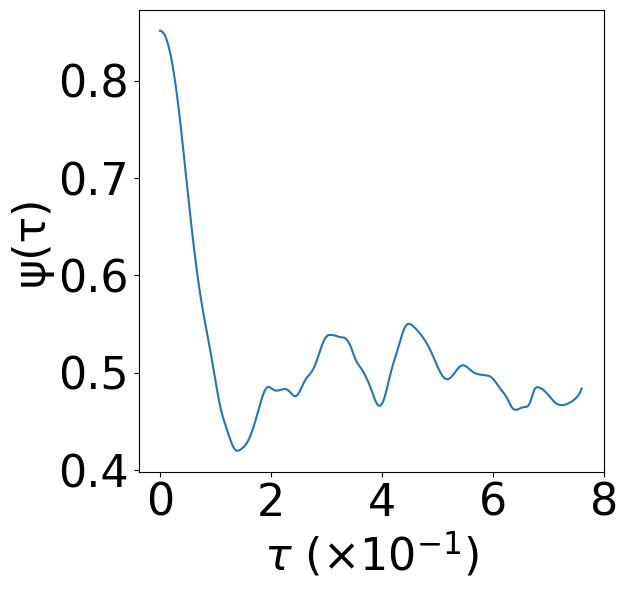}
\label{lastfm_omega10}
}%
\subfigure[$\omega_0=30$]{
\centering
\includegraphics[width=1.8in]{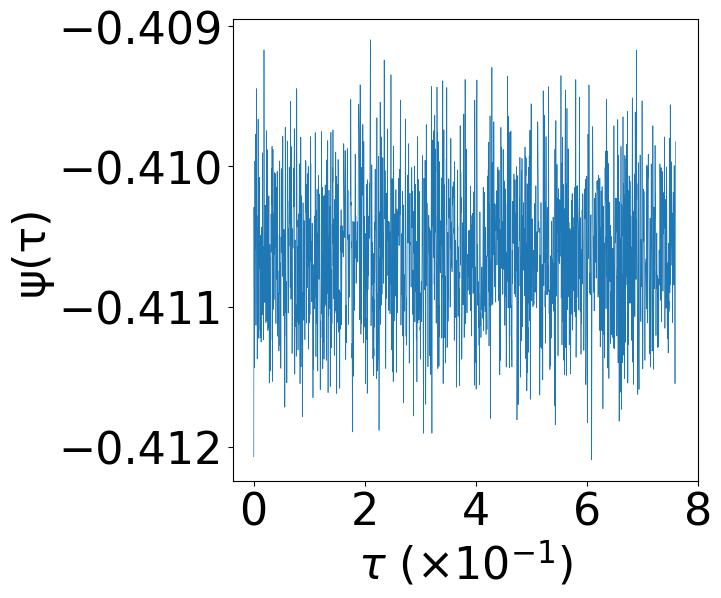}
\label{lastfm_omega30}
}%

\subfigure[$\omega_0=50$]{
\centering
\includegraphics[width=1.7in]{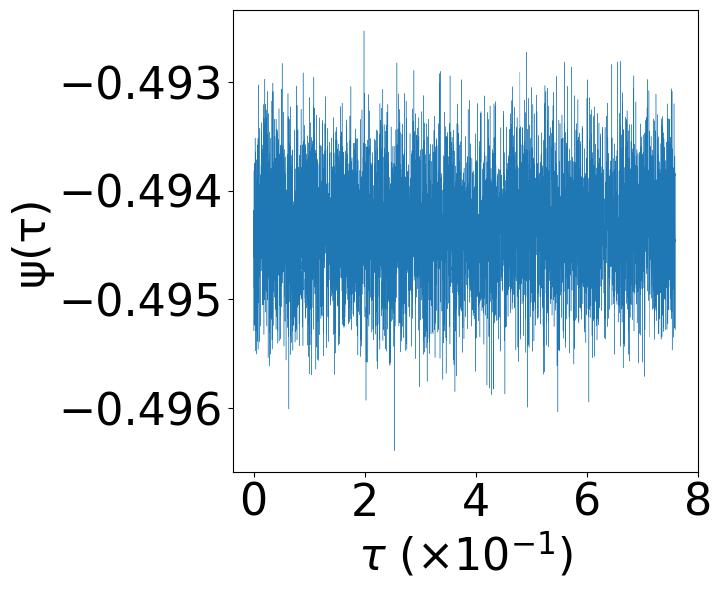}
\label{lastfm_omega50}
}%
\subfigure[$\omega_0=100$]{
\centering
\includegraphics[width=1.7in]{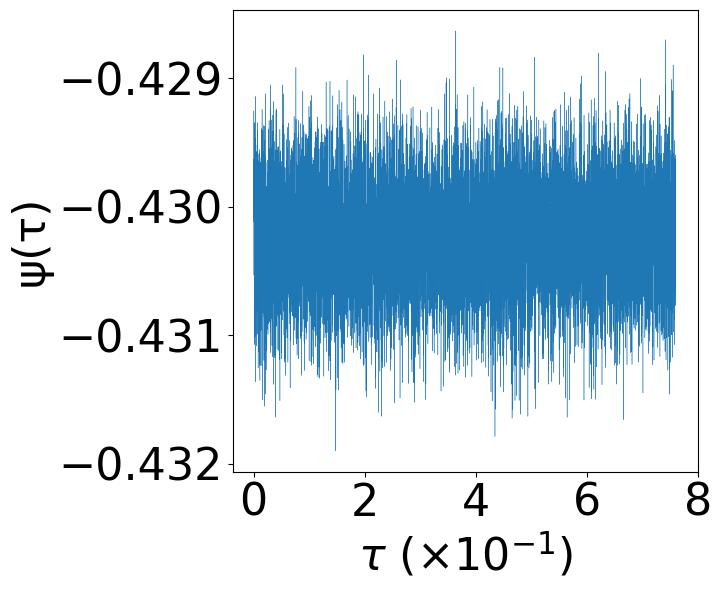}
\label{lastfm_omega100}
}%
\subfigure[$\omega_0=500$]{
\centering
\includegraphics[width=1.7in]{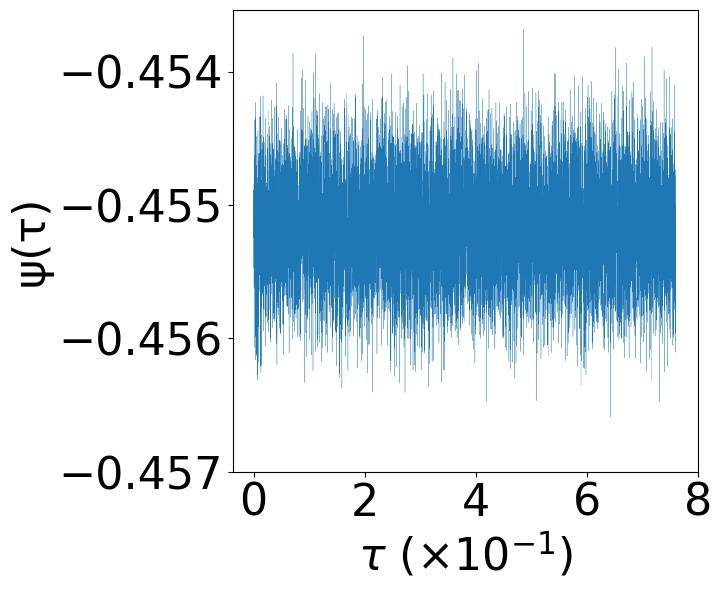}
\label{lastfm_omega500}
}%

    \caption{Visualization of kernels trained on LastFM with different values of $\omega_0$. The horizontal axis stands for the relative time position $\tau$ and the vertical axis stands for kernel weight $\psi(\tau)$ at the corresponding point (see eq.\mbox{\ref{eq:omega_tau}}). }
    \label{fig:lastfm_kernel}
\end{figure}

\begin{figure}[h!]
	\centering

\subfigure[$\omega_0=0.005$]{
\centering
\includegraphics[width=1.8in]{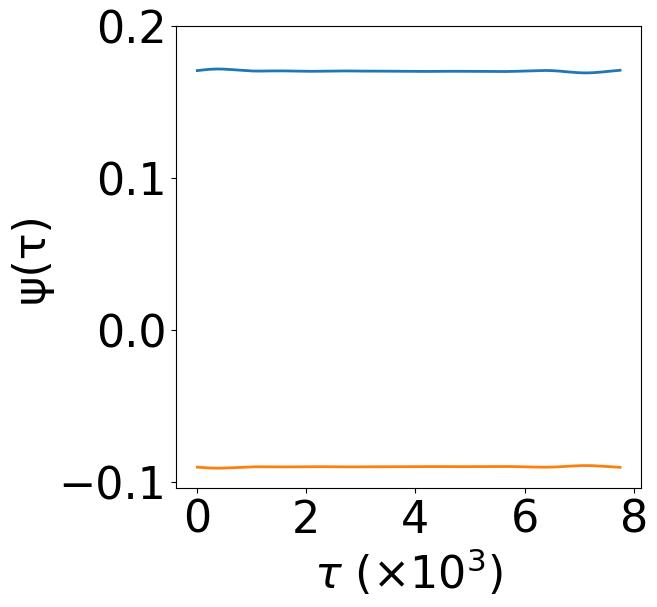}
\label{retweet_omega5e-3}
}%
\subfigure[$\omega_0=0.01$]{
\centering
\includegraphics[width=1.8in]{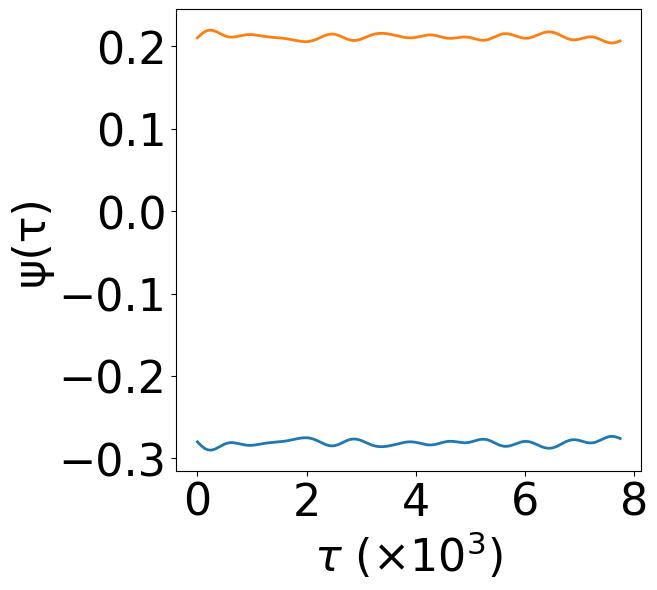}
\label{retweet_omega1e-2}
}%
\subfigure[$\omega_0=0.1$]{
\centering
\includegraphics[width=1.8in]{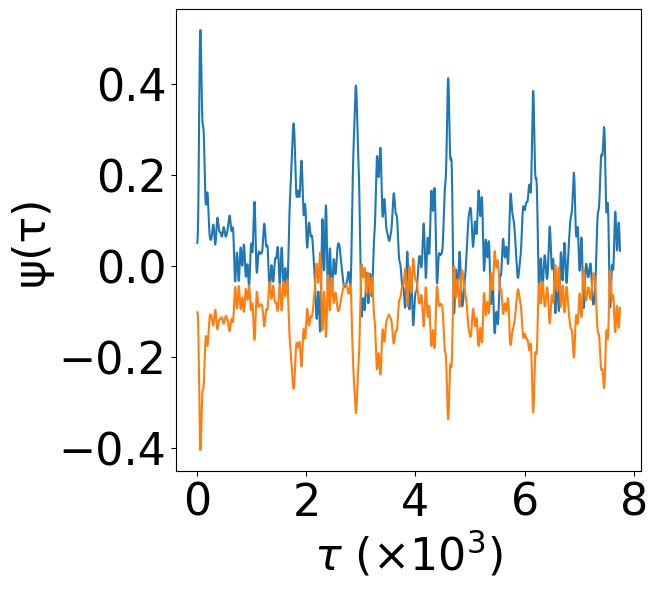}
\label{retweet_omega1e-1}
}%

\subfigure[$\omega_0=1$]{
\centering
\includegraphics[width=1.8in]{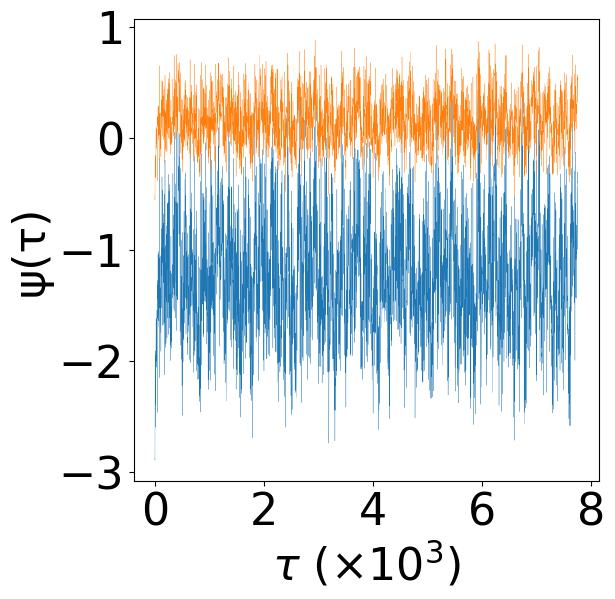}
\label{retweet_omega1}
}%
\subfigure[$\omega_0=5$]{
\centering
\includegraphics[width=1.8in]{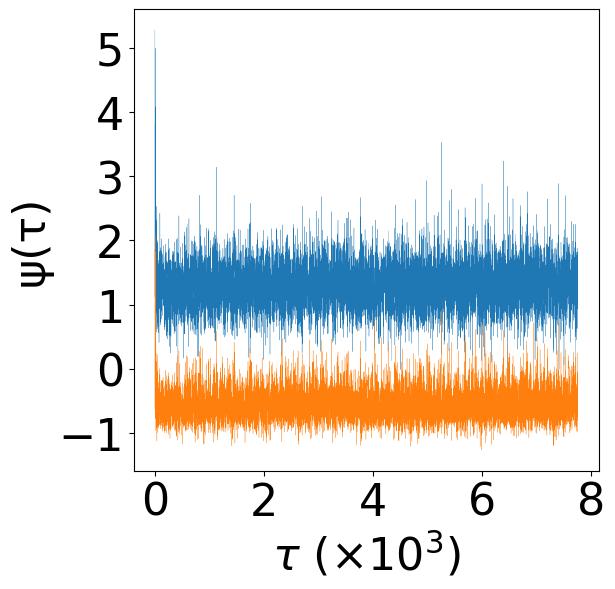}
\label{retweet_omega5}
}%
\subfigure[$\omega_0=10$]{
\centering
\includegraphics[width=1.9in]{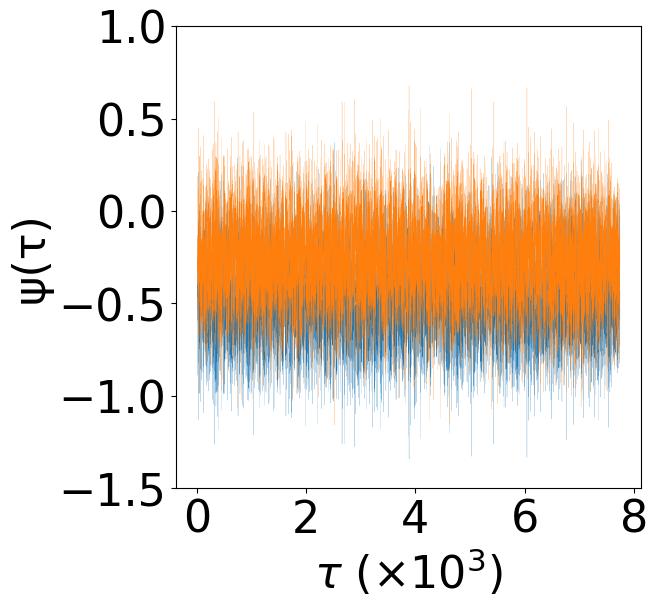}
\label{retweet_omega10}
}%

    \caption{Visualization of kernels trained on Retweet with different values of $\omega_0$. The horizontal axis stands for the relative time position $\tau$ and the vertical axis stands for kernel weight $\psi(\tau)$ at the corresponding point (see eq.\mbox{\ref{eq:omega_tau}}). Note that curves in different colours represent different channels.}
    \label{fig:retweet_kernel}
\end{figure}

According to \mbox{\citep{ckconv}}, the parameter $\omega_0$ is also a vital hyperparameter (see eq.~\mbox{\ref{eq:omega_tau}}), which governs the rate oscillation of the continuous kernel. 
For LastFM, a dataset of a small time scale ($\delta=0.76$), we tuned the value of $\omega_0$ within [1, 10, 30, 50, 100, 500], while other parameters are fixed to an optimal combination. Similarly, when training on Retweet ($\delta=2583$), $\omega_0$ is tuned within [0.005, 0.01, 0.1, 1, 5, 10] to handle a large time scale. The NLL results are shown in Fig.\mbox{\ref{fig:param2}}, and visualizations of the kernels trained with different $\omega_0$ values are displayed in Fig.\mbox{\ref{fig:lastfm_kernel}} and \mbox{\ref{fig:retweet_kernel}}. Among all the $\omega_0$ values tested, $\omega_0=10$ (Fig.\mbox{\ref{lastfm_omega10}}) leads to the best performance for LastFM. And for Retweet, $\omega_0=1$ yields relatively good results. As $\omega_0$ controls the oscillation of the kernel, a kernel with too small oscillation (e.g., Fig.\mbox{\ref{lastfm_omega1}} and  Fig.\mbox{\ref{retweet_omega5e-3}}) may lack flexibility, while one with too large oscillation (e.g., Fig.\mbox{\ref{lastfm_omega500} and Fig.\mbox{\ref{retweet_omega10}}}) could be oversensitive to time, making it uninterpretable and susceptible to overfitting. Thus, in order to find an optimal value of $omega_0$, careful tuning is required within an appropriate range.

\section{Conclusion}
\label{conclusion}
In this work, we propose CTPP, a TPP learning framework, which is capable of enhancing the modelling performance by incorporating local event contexts. In order to be better aware of the local context of a certain event, a continuous-time convolutional network is applied, together with an RNN. The proposed architecture is able to capture and fuse global and local temporal dependency, thus enhancing modelling performance. Through our comparison experiments on three real-world datasets against state-of-the-art methods, the superiority of our proposed method is validated. For future work, we are going to develop a better approach to model the temporal dynamics and uncertainties of the latent states, since time intervals are passed  to the model as simple scalars and the latent states are all deterministic, lacking robustness against noise and prone to overfitting.

\section*{Acknowledgements}
This work was supported by the Natural Science Foundation of China under Grant 62276053 and the Sichuan Science and Technology Program (Nos. 2021YFG0018, 2022YFG0038).



\bibliographystyle{plainnat} 

\bibliography{bibitems}

\begin{thebibliography}{49}
\providecommand{\natexlab}[1]{#1}
\providecommand{\url}[1]{\texttt{#1}}
\expandafter\ifx\csname urlstyle\endcsname\relax
  \providecommand{\doi}[1]{doi: #1}\else
  \providecommand{\doi}{doi: \begingroup \urlstyle{rm}\Url}\fi

\bibitem[Bacry et~al.(2015)Bacry, Mastromatteo, and Muzy]{finance}
Emmanuel Bacry, Iacopo Mastromatteo, and Jean-Fran{\c{c}}ois Muzy.
\newblock Hawkes processes in finance.
\newblock \emph{Market Microstructure and Liquidity}, 1\penalty0 (01):\penalty0
  1550005, 2015.

\bibitem[Bai et~al.(2018)Bai, Kolter, and Koltun]{TCN}
Shaojie Bai, J~Zico Kolter, and Vladlen Koltun.
\newblock An empirical evaluation of generic convolutional and recurrent
  networks for sequence modeling.
\newblock \emph{arXiv preprint arXiv:1803.01271}, 2018.

\bibitem[Bai et~al.(2019)Bai, Zou, Zhao, Du, Liu, Nie, and Wen]{CTRec}
Ting Bai, Lixin Zou, Wayne~Xin Zhao, Pan Du, Weidong Liu, Jian-Yun Nie, and
  Ji-Rong Wen.
\newblock Ctrec: A long-short demands evolution model for continuous-time
  recommendation.
\newblock In \emph{Proceedings of the 42nd International ACM SIGIR Conference
  on Research and Development in Information Retrieval}, pages 675--684, 2019.

\bibitem[Bertin-Mahieux et~al.(2011)Bertin-Mahieux, Ellis, Whitman, and
  Lamere]{lastfm}
Thierry Bertin-Mahieux, Daniel~P.W. Ellis, Brian Whitman, and Paul Lamere.
\newblock The million song dataset.
\newblock In \emph{{Proceedings of the 12th International Conference on Music
  Information Retrieval ({ISMIR} 2011)}}, 2011.

\bibitem[Chung et~al.(2014)Chung, Gulcehre, Cho, and
  Bengio]{chung2014empirical}
Junyoung Chung, Caglar Gulcehre, Kyunghyun Cho, and Yoshua Bengio.
\newblock Empirical evaluation of gated recurrent neural networks on sequence
  modeling.
\newblock In \emph{NIPS 2014 Workshop on Deep Learning, December 2014}, 2014.

\bibitem[Cui et~al.(2023)Cui, Sun, Pan, Liu, and Xu]{HawkesRecommend}
Zhihong Cui, Xiangguo Sun, Li~Pan, Shijun Liu, and Guandong Xu.
\newblock Event-based incremental recommendation via factors mixed hawkes
  process.
\newblock \emph{Information Sciences}, 639:\penalty0 119007, 2023.
\newblock ISSN 0020-0255.

\bibitem[Daley and Vere-Jones(2008)]{pointprocess}
Daryl~J Daley and David Vere-Jones.
\newblock \emph{An Introduction to the Theory of Point Processes. Volume II:
  General Theory and Structure}.
\newblock Springer, 2008.

\bibitem[Dauphin et~al.(2017)Dauphin, Fan, Auli, and Grangier]{conv3}
Yann~N Dauphin, Angela Fan, Michael Auli, and David Grangier.
\newblock Language modeling with gated convolutional networks.
\newblock In \emph{International conference on machine learning}, pages
  933--941. PMLR, 2017.

\bibitem[Deng et~al.(2019)Deng, Rangwala, and Ning]{dynamicsocial}
Songgaojun Deng, Huzefa Rangwala, and Yue Ning.
\newblock Learning dynamic context graphs for predicting social events.
\newblock In \emph{Proceedings of the 25th ACM SIGKDD International Conference
  on Knowledge Discovery \& Data Mining}, pages 1007--1016, 2019.

\bibitem[Deng et~al.(2020)Deng, Rangwala, and Ning]{dynamicknowledge}
Songgaojun Deng, Huzefa Rangwala, and Yue Ning.
\newblock Dynamic knowledge graph based multi-event forecasting.
\newblock In \emph{Proceedings of the 26th ACM SIGKDD International Conference
  on Knowledge Discovery \& Data Mining}, pages 1585--1595, 2020.

\bibitem[Du et~al.(2016)Du, Dai, Trivedi, Upadhyay, Gomez-Rodriguez, and
  Song]{rmtpp}
Nan Du, Hanjun Dai, Rakshit Trivedi, Utkarsh Upadhyay, Manuel Gomez-Rodriguez,
  and Le~Song.
\newblock Recurrent marked temporal point processes: Embedding event history to
  vector.
\newblock In \emph{Proceedings of the 22nd ACM SIGKDD international conference
  on knowledge discovery and data mining}, pages 1555--1564, 2016.

\bibitem[Galkina and Grafeeva(2019)]{earthquake}
Alyona Galkina and Natalia Grafeeva.
\newblock Machine learning methods for earthquake prediction: A survey.
\newblock In \emph{Proceedings of the Fourth Conference on Software Engineering
  and Information Management (SEIM-2019), Saint Petersburg, Russia}, volume~13,
  page~25, 2019.

\bibitem[Gehring et~al.(2017{\natexlab{a}})Gehring, Auli, Grangier, and
  Dauphin]{conv1}
Jonas Gehring, Michael Auli, David Grangier, and Yann Dauphin.
\newblock A convolutional encoder model for neural machine translation.
\newblock In \emph{Proceedings of the 55th Annual Meeting of the Association
  for Computational Linguistics (Volume 1: Long Papers)}, pages 123--135,
  2017{\natexlab{a}}.

\bibitem[Gehring et~al.(2017{\natexlab{b}})Gehring, Auli, Grangier, Yarats, and
  Dauphin]{conv2}
Jonas Gehring, Michael Auli, David Grangier, Denis Yarats, and Yann~N Dauphin.
\newblock Convolutional sequence to sequence learning.
\newblock In \emph{International conference on machine learning}, pages
  1243--1252. PMLR, 2017{\natexlab{b}}.

\bibitem[Gulati et~al.(2020)Gulati, Qin, Chiu, Parmar, Zhang, Yu, Han, Wang,
  Zhang, Wu, and Pang]{speech}
Anmol Gulati, James Qin, Chung-Cheng Chiu, Niki Parmar, Yu~Zhang, Jiahui Yu,
  Wei Han, Shibo Wang, Zhengdong Zhang, Yonghui Wu, and Ruoming Pang.
\newblock Conformer: Convolution-augmented transformer for speech recognition.
\newblock In \emph{Proceedings of the Annual Conference of the International
  Speech Communication Association}, pages 5036--5040, 2020.

\bibitem[Hawkes(1971)]{Hawkes}
Alan~G Hawkes.
\newblock Point spectra of some mutually exciting point processes.
\newblock \emph{Journal of the Royal Statistical Society: Series B
  (Methodological)}, 33\penalty0 (3):\penalty0 438--443, 1971.

\bibitem[Kapoor et~al.(2014)Kapoor, Sun, Srivastava, and Ye]{eg1}
Komal Kapoor, Mingxuan Sun, Jaideep Srivastava, and Tao Ye.
\newblock A hazard based approach to user return time prediction.
\newblock In \emph{Proceedings of the 20th ACM SIGKDD international conference
  on Knowledge discovery and data mining}, pages 1719--1728, 2014.

\bibitem[Kapoor et~al.(2015)Kapoor, Subbian, Srivastava, and Schrater]{eg2}
Komal Kapoor, Karthik Subbian, Jaideep Srivastava, and Paul Schrater.
\newblock Just in time recommendations: Modeling the dynamics of boredom in
  activity streams.
\newblock In \emph{Proceedings of the eighth ACM international conference on
  web search and data mining}, pages 233--242, 2015.

\bibitem[Kenton and Toutanova(2019)]{bert}
Jacob Devlin Ming-Wei~Chang Kenton and Lee~Kristina Toutanova.
\newblock Bert: Pre-training of deep bidirectional transformers for language
  understanding.
\newblock In \emph{Proceedings of NAACL-HLT}, pages 4171--4186, 2019.

\bibitem[Kingma and Ba(2015)]{Adam}
Diederik~P Kingma and Jimmy Ba.
\newblock Adam: A method for stochastic optimization.
\newblock In \emph{International Conference on Learning Representations}, 2015.

\bibitem[LeCun et~al.(1989)LeCun, Boser, Denker, Henderson, Howard, Hubbard,
  and Jackel]{convolution}
Yann LeCun, Bernhard Boser, John~S Denker, Donnie Henderson, Richard~E Howard,
  Wayne Hubbard, and Lawrence~D Jackel.
\newblock Backpropagation applied to handwritten zip code recognition.
\newblock \emph{Neural computation}, 1\penalty0 (4):\penalty0 541--551, 1989.

\bibitem[Leskovec and Krevl(2014)]{stackoverflow}
Jure Leskovec and Andrej Krevl.
\newblock {SNAP Datasets}: {Stanford} large network dataset collection.
\newblock \url{http://snap.stanford.edu/data}, June 2014.

\bibitem[Lin et~al.(2022)Lin, Wu, Zhao, Pai, and Li]{generative2022}
Haitao Lin, Lirong Wu, Guojiang Zhao, Liu Pai, and Stan~Z. Li.
\newblock Exploring generative neural temporal point process.
\newblock \emph{Transactions on Machine Learning Research}, 2022.
\newblock ISSN 2835-8856.

\bibitem[Liu et~al.(2023{\natexlab{a}})Liu, Tian, Kang, and Wan]{anomaly}
Liang Liu, Ling Tian, Zhao Kang, and Tianqi Wan.
\newblock Spacecraft anomaly detection with attention temporal convolution
  networks.
\newblock \emph{Neural Computing and Applications}, pages 1--9,
  2023{\natexlab{a}}.

\bibitem[Liu et~al.(2023{\natexlab{b}})Liu, Zhang, Lu, Chen, and Wei]{traffic}
Wei Liu, Tao Zhang, Yisheng Lu, Jun Chen, and Longsheng Wei.
\newblock That-net: Two-layer hidden state aggregation based two-stream network
  for traffic accident prediction.
\newblock \emph{Information Sciences}, 634:\penalty0 744--760,
  2023{\natexlab{b}}.
\newblock ISSN 0020-0255.

\bibitem[Ma et~al.(2019)Ma, Yu, Tian, Chen, and Ng]{globallocal1}
Qianli Ma, Liuhong Yu, Shuai Tian, Enhuan Chen, and Wing W.~Y. Ng.
\newblock Global-local mutual attention model for text classification.
\newblock \emph{IEEE/ACM Transactions on Audio, Speech, and Language
  Processing}, 27\penalty0 (12):\penalty0 2127--2139, 2019.

\bibitem[Mei and Eisner(2017)]{nhp}
Hongyuan Mei and Jason~M Eisner.
\newblock The neural hawkes process: A neurally self-modulating multivariate
  point process.
\newblock \emph{Advances in neural information processing systems}, 30, 2017.

\bibitem[Omi et~al.(2019)Omi, Aihara, et~al.]{FullyNN}
Takahiro Omi, Kazuyuki Aihara, et~al.
\newblock Fully neural network based model for general temporal point
  processes.
\newblock \emph{Advances in neural information processing systems}, 32, 2019.

\bibitem[Pan et~al.(2020)Pan, Huang, Lian, and Chen]{generative2020}
Zhen Pan, Zhenya Huang, Defu Lian, and Enhong Chen.
\newblock A variational point process model for social event sequences.
\newblock In \emph{Proceedings of the AAAI Conference on Artificial
  Intelligence}, volume~34, pages 173--180, 2020.

\bibitem[Reinhart(2018)]{self-exciting}
Alex Reinhart.
\newblock A review of self-exciting spatio-temporal point processes and their
  applications.
\newblock \emph{Statistical Science}, 33\penalty0 (3):\penalty0 299--318, 2018.

\bibitem[Romero et~al.(2021)Romero, Kuzina, Bekkers, Tomczak, and
  Hoogendoorn]{ckconv}
David~W Romero, Anna Kuzina, Erik~J Bekkers, Jakub~Mikolaj Tomczak, and Mark
  Hoogendoorn.
\newblock Ckconv: Continuous kernel convolution for sequential data.
\newblock In \emph{International Conference on Learning Representations}, 2021.

\bibitem[Rosenfeld(2000)]{ngram1}
Ronald Rosenfeld.
\newblock Two decades of statistical language modeling: Where do we go from
  here?
\newblock \emph{Proceedings of the IEEE}, 88\penalty0 (8):\penalty0 1270--1278,
  2000.

\bibitem[Sen et~al.(2019)Sen, Yu, and Dhillon]{global_local}
Rajat Sen, Hsiang-Fu Yu, and Inderjit~S Dhillon.
\newblock Think globally, act locally: A deep neural network approach to
  high-dimensional time series forecasting.
\newblock \emph{Advances in neural information processing systems}, 32, 2019.

\bibitem[Shchur et~al.(2019)Shchur, Bilo{\v{s}}, and G{\"u}nnemann]{lognormmix}
Oleksandr Shchur, Marin Bilo{\v{s}}, and Stephan G{\"u}nnemann.
\newblock Intensity-free learning of temporal point processes.
\newblock In \emph{International Conference on Learning Representations}, 2019.

\bibitem[Sitzmann et~al.(2020)Sitzmann, Martel, Bergman, Lindell, and
  Wetzstein]{siren}
Vincent Sitzmann, Julien Martel, Alexander Bergman, David Lindell, and Gordon
  Wetzstein.
\newblock Implicit neural representations with periodic activation functions.
\newblock \emph{Advances in Neural Information Processing Systems},
  33:\penalty0 7462--7473, 2020.

\bibitem[Soen et~al.(2021)Soen, Mathews, Grixti-Cheng, and Xie]{unipoint}
Alexander Soen, Alexander Mathews, Daniel Grixti-Cheng, and Lexing Xie.
\newblock Unipoint: Universally approximating point processes intensities.
\newblock In \emph{Proceedings of the AAAI Conference on Artificial
  Intelligence}, volume~35, pages 9685--9694, 2021.

\bibitem[Tran et~al.(2022)Tran, Wei, Ruan, McGowan, Susanj, and
  Strimel]{globallocal2}
Thanh Tran, Kai Wei, Weitong Ruan, Ross McGowan, Nathan Susanj, and Grant~P
  Strimel.
\newblock Adaptive global-local context fusion for multi-turn spoken language
  understanding.
\newblock In \emph{Proceedings of the AAAI Conference on Artificial
  Intelligence}, volume~36, pages 12622--12628, 2022.

\bibitem[van~den Oord et~al.(2016)van~den Oord, Dieleman, Zen, Simonyan,
  Vinyals, Graves, Kalchbrenner, Senior, and Kavukcuoglu]{wavenet}
A{\"a}ron van~den Oord, Sander Dieleman, Heiga Zen, Karen Simonyan, Oriol
  Vinyals, Alex Graves, Nal Kalchbrenner, Andrew Senior, and Koray Kavukcuoglu.
\newblock Wavenet: A generative model for raw audio.
\newblock In \emph{9th ISCA Speech Synthesis Workshop}, pages 125--125, 2016.

\bibitem[Vaswani et~al.(2017)Vaswani, Shazeer, Parmar, Uszkoreit, Jones, Gomez,
  Kaiser, and Polosukhin]{Transformer}
Ashish Vaswani, Noam Shazeer, Niki Parmar, Jakob Uszkoreit, Llion Jones,
  Aidan~N Gomez, {\L}ukasz Kaiser, and Illia Polosukhin.
\newblock Attention is all you need.
\newblock \emph{Advances in neural information processing systems}, 30, 2017.

\bibitem[Wu et~al.(2019)Wu, Li, Zhao, and Qian]{POI}
Yuxia Wu, Ke~Li, Guoshuai Zhao, and Xueming Qian.
\newblock Long- and short-term preference learning for next poi recommendation.
\newblock In \emph{Proceedings of the 28th ACM International Conference on
  Information and Knowledge Management}, page 2301–2304, New York, NY, USA,
  2019. Association for Computing Machinery.

\bibitem[Xiao et~al.(2017)Xiao, Yan, Yang, Zha, and Chu]{lstmtpp}
Shuai Xiao, Junchi Yan, Xiaokang Yang, Hongyuan Zha, and Stephen Chu.
\newblock Modeling the intensity function of point process via recurrent neural
  networks.
\newblock In \emph{Proceedings of the AAAI Conference on Artificial
  Intelligence}, volume~31, 2017.

\bibitem[Yang et~al.(2022)Yang, Mei, and Eisner]{2022transformer}
Chenghao Yang, Hongyuan Mei, and Jason Eisner.
\newblock Transformer embeddings of irregularly spaced events and their
  participants.
\newblock In \emph{Proceedings of the Tenth International Conference on
  Learning Representations}, 2022.

\bibitem[Yuan et~al.(2020)Yuan, Wang, Yu, Liu, and Li]{YUAN2020122}
Weihua Yuan, Hong Wang, Xiaomei Yu, Nan Liu, and Zhenghao Li.
\newblock Attention-based context-aware sequential recommendation model.
\newblock \emph{Information Sciences}, 510:\penalty0 122--134, 2020.
\newblock ISSN 0020-0255.

\bibitem[Zhang et~al.(2020)Zhang, Lipani, Kirnap, and Yilmaz]{SAHP}
Qiang Zhang, Aldo Lipani, Omer Kirnap, and Emine Yilmaz.
\newblock Self-attentive hawkes process.
\newblock In \emph{International conference on machine learning}, pages
  11183--11193. PMLR, 2020.

\bibitem[Zhao et~al.(2015{\natexlab{a}})Zhao, Sun, Ye, Chen, Lu, and
  Ramakrishnan]{multitask}
Liang Zhao, Qian Sun, Jieping Ye, Feng Chen, Chang-Tien Lu, and Naren
  Ramakrishnan.
\newblock Multi-task learning for spatio-temporal event forecasting.
\newblock In \emph{Proceedings of the 21th ACM SIGKDD international conference
  on knowledge discovery and data mining}, pages 1503--1512,
  2015{\natexlab{a}}.

\bibitem[Zhao et~al.(2015{\natexlab{b}})Zhao, Erdogdu, He, Rajaraman, and
  Leskovec]{retweet}
Qingyuan Zhao, Murat~A Erdogdu, Hera~Y He, Anand Rajaraman, and Jure Leskovec.
\newblock Seismic: A self-exciting point process model for predicting tweet
  popularity.
\newblock In \emph{Proceedings of the 21th ACM SIGKDD international conference
  on knowledge discovery and data mining}, pages 1513--1522,
  2015{\natexlab{b}}.

\bibitem[Zhu et~al.(2021)Zhu, Zhang, Ding, and Xie]{Fourier}
Shixiang Zhu, Minghe Zhang, Ruyi Ding, and Yao Xie.
\newblock Deep fourier kernel for self-attentive point processes.
\newblock In \emph{International Conference on Artificial Intelligence and
  Statistics}, pages 856--864. PMLR, 2021.

\bibitem[Zhu et~al.(2022)Zhu, Tang, Wang, Li, Guo, and Dietze]{ZHU2022}
Xiaofei Zhu, Gu~Tang, Pengfei Wang, Chenliang Li, Jiafeng Guo, and Stefan
  Dietze.
\newblock Dynamic global structure enhanced multi-channel graph neural network
  for session-based recommendation.
\newblock \emph{Information Sciences}, 2022.
\newblock ISSN 0020-0255.

\bibitem[Zuo et~al.(2020)Zuo, Jiang, Li, Zhao, and Zha]{THP}
Simiao Zuo, Haoming Jiang, Zichong Li, Tuo Zhao, and Hongyuan Zha.
\newblock Transformer hawkes process.
\newblock In \emph{International conference on machine learning}, pages
  11692--11702. PMLR, 2020.

\end{thebibliography}

\newpage
\par\noindent 
\parbox[t]{\linewidth}{
\noindent\parpic{\includegraphics[height=1.5in,width=1in,clip,keepaspectratio]{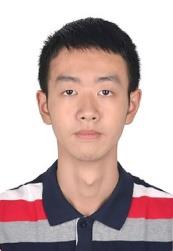}}
\noindent {\bf Wang-Tao Zhou}\
was born in Sichuan in 1997. He received his B.E. degree in Computer Science
and Technology from University of Electronic Science and Technology of China (UESTC) in 2020.
He is currently pursuing Ph.D. degree in the Department of Computer Science and Engineering,
UESTC. His research interest includes event forecasting, temporal point processes, and time series forecasting.}
\vspace{8\baselineskip}

\par\noindent 
\parbox[t]{\linewidth}{
\noindent\parpic{\includegraphics[height=1.5in,width=1in,clip,keepaspectratio]{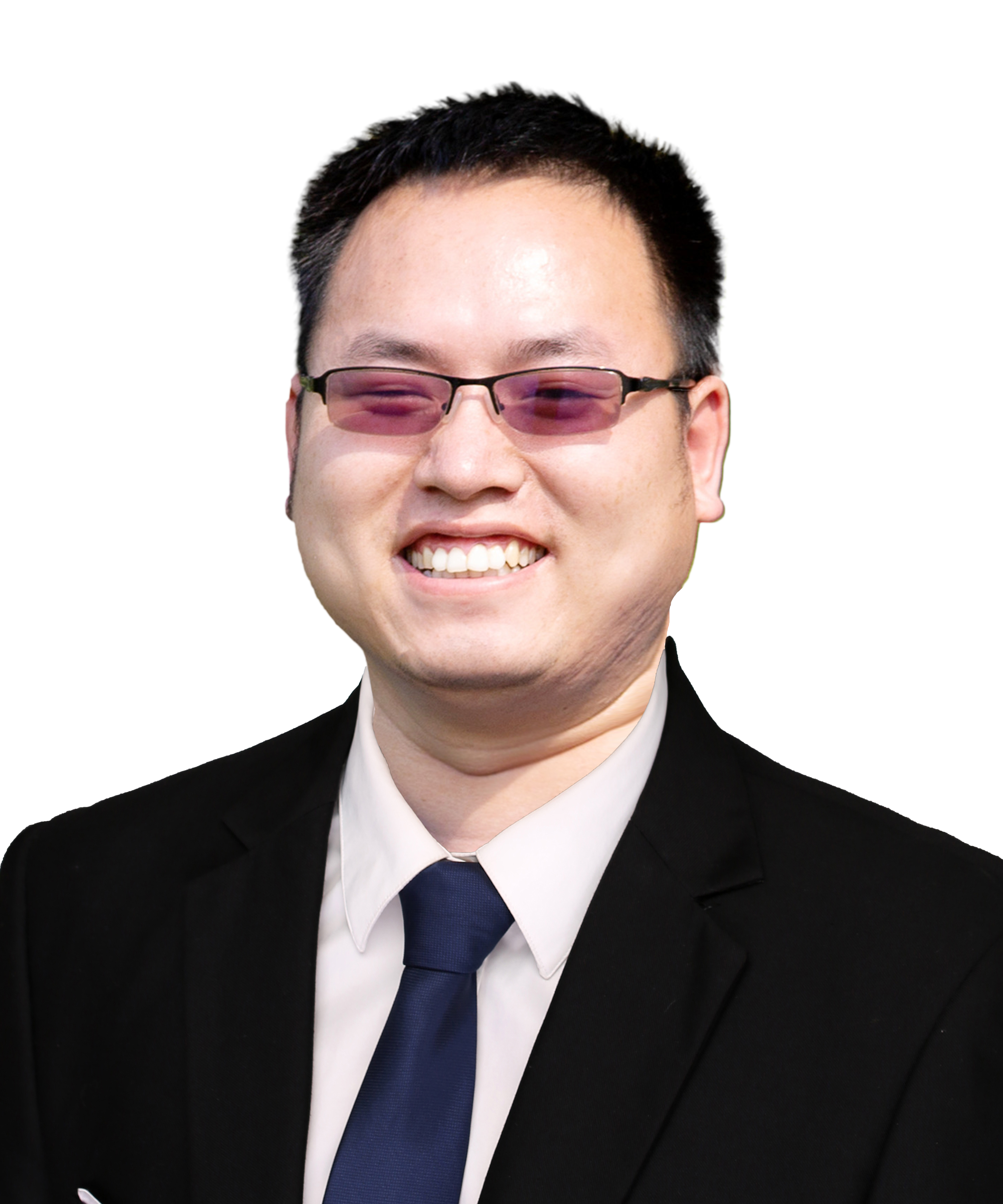}}
\noindent {\bf Zhao Kang}\
received the Ph.D. degree in computer science from Southern Illinois University Carbondale, Carbondale, IL, USA, in 2017. He is currently an Associate Professor at the School of Computer Science and Engineering, University of Electronic Science and Technology of China, Chengdu, China. He has published over 90 research papers in top-tier conferences and journals, including \textit{ICML}, \textit{NeurIPS}, \textit{AAAI}, \textit{IJCAI}, \textit{\textit{ICDE}, \textit{CVPR}, \textit{SIGKDD}, \textit{ECCV}, \textit{IEEE Transactions on Cybernetics}, \textit{IEEE Transactions on Image Processing}, \textit{IEEE Transactions on Knowledge and Data Engineering}, and \textit{IEEE Transactions on Neural Networks and Learning Systems}. His research interests are machine learning, pattern recognition, and data mining. Dr. Kang has been an AC/SPC/PC Member or a Reviewer for a number of top conferences, such as \textit{NeurIPS}, \textit{ICLR}, \textit{AAAI}, \textit{IJCAI}, \textit{CVPR}, \textit{ICCV}, \textit{ICML}, and \textit{ECCV}. He regularly serves as a Reviewer for \textit{the Journal of Machine Learning Research}, \textit{IEEE TRANSACTIONS ON PATTERN ANALYSIS AND MACHINE INTELLIGENCE}, \textit{IEEE TRANSACTIONS ON NEURAL NETWORKS AND LEARNING SYSTEMS}, \textit{IEEE TRANSACTIONS ON CYBERNETICS}, IEEE TRANSACTIONS ON KNOWLEDGE AND DATA ENGINEERING}, and \textit{IEEE TRANSACTIONS ON MULTIMEDIA}.}
\vspace{8\baselineskip}

\par\noindent 
\parbox[t]{\linewidth}{
\noindent\parpic{\includegraphics[height=1.5in,width=1in,clip,keepaspectratio]{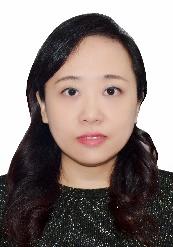}}
\noindent {\bf Ling Tian}\
received the B.S., M.S., and Ph.D. degrees from the School of Computer Science and
Engineering, University of Electronic Science and Technology of China (UESTC) in 2003, 2006
and 2010, respectively. She is currently a Professor at UESTC. She has edited 2 books and holds
over 30 China patents. She has contributed over 10 technology proposals to the standardizations
such as China Audio and Video Standard (AVS) and China Cloud Computing standard. Her
research interests include image/video coding, knowledge graph and knowledge reasoning.}
\vspace{8\baselineskip}

\par\noindent 
\parbox[t]{\linewidth}{
\noindent\parpic{\includegraphics[height=1.5in,width=1in,clip,keepaspectratio]{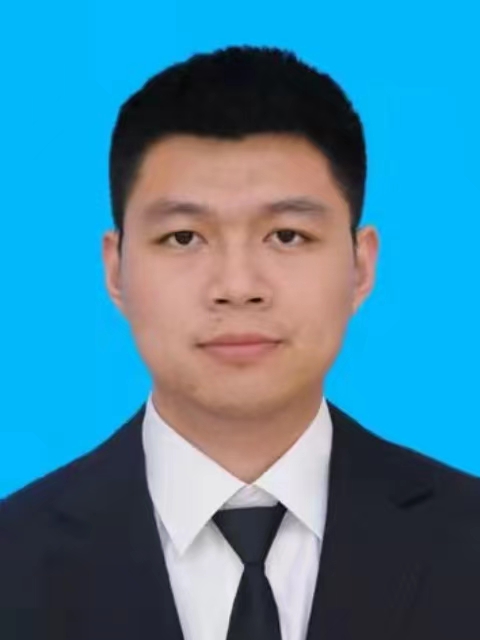}}
\noindent {\bf Yi Su}\
was born in Beijing in 1986. He received a B.E. degree in Electronic Information Engineering from Northwestern Polytechnical University(NPU) in 2009 and an M.S. degree in Electromagnetic Field and Microwave Technology from NPU in 2012. He is currently pursuing a Ph.D. degree in the Department of Computer Science and Engineering, University of electronic science and Technology of China (UESTC). His research interest includes Object detection, transfer learning, and deep learning.}
\vspace{8\baselineskip}

\end{document}